\documentclass{article}

\usepackage{arxiv}

\usepackage{xspace}
\usepackage[numbers]{natbib}
\usepackage{graphicx}
\usepackage{diagbox}
\usepackage{multirow}
\usepackage[nopostdot,nogroupskip,nonumberlist]{glossaries}
\usepackage{afterpage}
\usepackage{tabularx}
\usepackage{xltabular}
\usepackage{booktabs}
\usepackage{longtable}
\usepackage{forest}
\usepackage{lscape}
\usepackage{adjustbox}
\PassOptionsToPackage{hyphens}{url}\usepackage{hyperref}  

\makeatletter
\def\@BTrule[#1]{%
  \ifx\longtable\undefined
    \let\@BTswitch\@BTnormal
  \else\ifx\hline\LT@hline
    \nobreak
    \let\@BTswitch\@BLTrule
  \else
     \let\@BTswitch\@BTnormal
  \fi\fi
  \global\@thisrulewidth=#1\relax
  \ifnum\@thisruleclass=\tw@\vskip\@aboverulesep\else
  \ifnum\@lastruleclass=\z@\vskip\@aboverulesep\else
  \ifnum\@lastruleclass=\@ne\vskip\doublerulesep\fi\fi\fi
  \@BTswitch}
\makeatother

\AtBeginDocument{%
  \providecommand\BibTeX{{%
    \normalfont B\kern-0.5em{\scshape i\kern-0.25em b}\kern-0.8em\TeX}}}

\newif\iffinal
\finaltrue

\iffinal
  \newcommand\christian[1]{}
  \newcommand\simon[1]{}
\else
  \newcommand\christian[1]{{\color{red}[***Christian: #1]\\}}
  \newcommand\simon[1]{{\color{blue}[***Simon: #1]\\}}
\fi

\begin{document}

\newcommand\tightenfig{\vspace{0pt}}
\newcommand\tightentab{\vspace{0pt}}

\title{Fairness in Machine Learning: A Survey}

\author{Simon Caton \\
University College Dublin \\
Dublin, Ireland\\
simon.caton@ucd.ie
\And 
Christian Haas \\
University of Nebraska at Omaha\\
Omaha, US\\
christianhaas@unomaha.edu}

\newacronym{ml}{ML}{Machine Learning}
\newcommand{\ml}{\gls{ml}\xspace}

\newacronym{tpr}{TPR}{True Positive Rate}
\newcommand{\tpr}{\gls{tpr}\xspace}

\newacronym{fpr}{FPR}{False Positive Rate}
\newcommand{\fpr}{\gls{fpr}\xspace}

\newacronym{fnr}{FNR}{False Negative Rate}
\newcommand{\fnr}{\gls{fnr}\xspace}

\newacronym{tnr}{TNR}{True Negative Rate}
\newcommand{\tnr}{\gls{tnr}\xspace}

\newacronym{gdpr}{GDPR}{General Data Protection Regulation}
\newcommand{\gdpr}{\gls{gdpr}\xspace}

\renewcommand{\sectionautorefname}{Section}

\newcommand{\projectName}{***REMOVE PROJECT NAME COMMAND***}
\maketitle
\begin{abstract}
As Machine Learning technologies become increasingly used in contexts that affect citizens, companies as well as researchers need to be confident that their application of these methods will not have unexpected social implications, such as bias towards gender, ethnicity, and/or people with disabilities. There is significant literature on approaches to mitigate bias and promote fairness, yet the area is complex and hard to penetrate for newcomers to the domain. This article seeks to provide an overview of the different schools of thought and approaches to mitigating (social) biases and increase fairness in the Machine Learning literature. It organises approaches into the widely accepted framework of pre-processing, in-processing, and post-processing methods, subcategorizing into a further 11 method areas. Although much of the literature emphasizes binary classification, a discussion of fairness in regression, recommender systems, unsupervised learning, and natural language processing is also provided along with a selection of currently available open source libraries. The article concludes by summarising open challenges articulated as four dilemmas for fairness research.
\end{abstract}



\keywords{fairness, accountability, transparency, machine learning}


\section{Introduction}
\ml technologies solve challenging problems which often have high social impact, such as examining re-offence rates (e.g. \cite{berk2018fairness, brennan2013emergence, ferguson2015big, oneil2016, angwin2016machine, berk2019accuracy}), automating chat and (tech) support, and screening job applications (see \cite{raghavan2020mitigating, van2019hiring}). 
Yet, approaches in \ml have ``found dark skin unattractive'',\footnote{\url{https://www.theguardian.com/technology/2016/sep/08/artificial-intelligence-beauty-contest-doesnt-like-black-people}} claimed that ``black people reoffend more'',\footnote{\url{https://www.propublica.org/article/machine-bias-risk-assessments-in-criminal-sentencing}} and created a Neo-Nazi sexbot.\footnote{\url{https://www.technologyreview.com/s/610634/microsofts-neo-nazi-sexbot-was-a-great-lesson-for-makers-of-ai-assistants/}} With the increasingly widespread use of automated decision making and \ml approaches in general, fairness considerations in \ml have gained significant attention in research and practice in the 2010s. However, from a historical perspective these modern approaches often build on prior definitions, concepts, and considerations that have been suggested and developed over the past five decades. Specifically, there is a rich set of fairness-related work in a variety of disciplines, often with concepts that are similar or equal to current \ml fairness research \cite{Hutchinson2019a}. For example, discrimination in hiring decisions has been examined since the 1960s \cite{guion1966employment}. 
Research into (un)fairness, discrimination, and bias emerged after the 1964 US Civil Rights act, making it illegal to discriminate based on certain criteria in the context of government agencies (Title VI) and employment (Title VII). Two initial foci of fairness research were unfairness of standardized tests in higher education/university contexts \cite{Cleary1966,Cleary1968} as well as discrimination in employment-based concepts \cite{guion1966employment}. The first years after the Civil Rights act saw the emergence of a variety of definitions, metrics, and scholarly disputes about the applicability of various definitions and fairness concepts as well as the realizations that some concepts (such as group-based vs individual notions of fairness) can be incompatible. 

When comparing current work in \ml with initial work in fairness, it is noteworthy that much of the early literature considers regression settings as well a correlation-based definition of fairness properties of an underlying mechanism due to the focus on test fairness with a continuous target variable (see e.g., \cite{darlington1971}). However, the general notions transfer to (binary) classification settings and thus define essential concepts such as protected/demographic variables (e.g. \cite{Cleary1966,Cleary1968}), notions of group vs individual fairness (e.g. \cite{Thorndike1971,Sawyer1976}), impossibility conditions between different fairness conditions (\cite{darlington1971}), and fairness quantification based on metrics (e.g., true positive rates, \cite{Cole1973}).

Despite the increased discussion of different aspects and viewpoints of fairness in the 1970s as well as the founding of many modern fairness concepts, no general consensus as to what constitutes fairness or if it can/should be quantified emerged based on this first wave of fairness research. As \cite{Hutchinson2019a} note, some of the related discussions resonate with current discussions in \ml, e.g., the difficulty that different notions can be incompatible with each other, or the fact that each specific quantified measurement of fairness seems to have particular downsides. 

More recently, many researchers (e.g. \cite{green2018, veale2018, burrell2016, corbett2018, barabas2018, binns2017, feldman2015, pedreshi2008, zwitter2014big, boyd2011six, boyd2012critical, metcalf2016human, lepri2017}), governments (e.g. the EU in \cite{AIManifesto, euTender}, and the US in \cite{podesta2014, munoz2016}), policies like the \gdpr, NGOs (e.g. the Association of Internet Researchers \cite{markham2012ethical}) and the media have fervently called for more societal accountability and social understanding of \ml . There is a recognition in the literature that often data is the problem, i.e. intrinsic biases in the sample will manifest themselves in any model built on the data \cite{boyd2012critical, binns2017}, inappropriate uses of data leading to (un)conscious bias(es) \cite{boyd2011six, boyd2012critical}, data veracity and quality \cite{zwitter2014big}, data relativity and context shifts \cite{boyd2012critical, rost2013representation, gonzalez2014assessing}, and subjectivity filters \cite{boyd2011six}. Even for skilled \ml researchers, the array of challenges can be overwhelming and current \ml libraries often do not yet accommodate means to ascertain social accountability. Note that data is not the only source of bias and discrimination, here we refer to \cite{mehrabi2019survey} for a general discussion on the main types of bias and discrimination in \ml . 

\begin{figure}[ht]
    \centering
    \includegraphics[width=.5\textwidth]{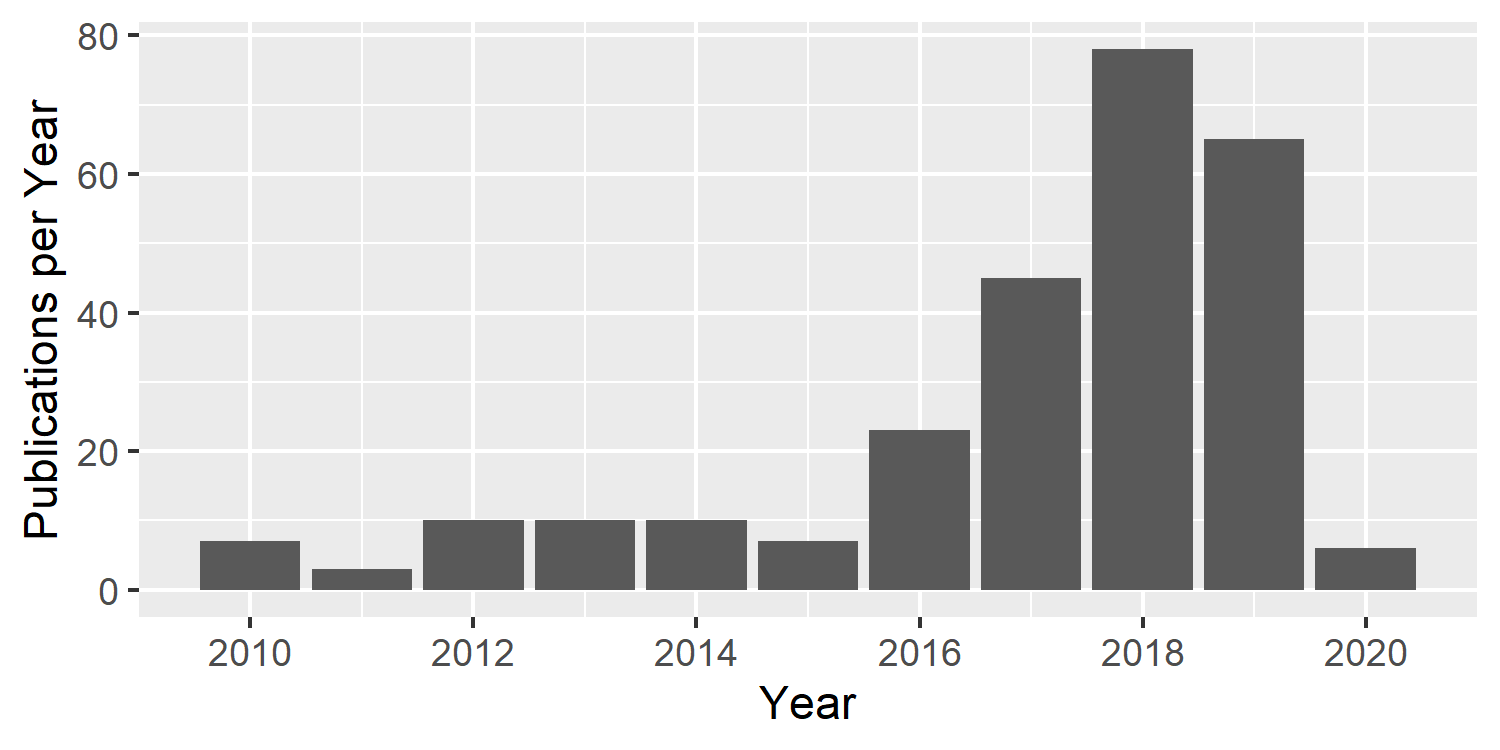}
    \caption{Number of Papers related to Fairness in \ml research based on cited articles in this survey.}
    \label{fig:numpapers}
    \tightenfig
\end{figure}

\ml researchers have responded to this call, developing a large number of metrics to quantify fairness in decisions (automated or otherwise) and mitigate any bias and unfairness issues in \ml . Figure \ref{fig:numpapers} shows the number of papers, starting in 2010, that have been published in the fairness in \ml domain. 
These numbers are based on the articles referenced in our survey. The figure shows a clear uptick in papers starting in 2016 and 2017.\footnote{The numbers for 2020 are incomplete as the survey was submitted during the year.} 

In this respect, this article aims to provide an entry-level overview of the current state of the art for fairness in \ml . This article builds on other similarly themed reviews that have focused on the history of fairness in \ml \cite{Hutchinson2019a}, a multidisciplinary survey of discrimination analysis \cite{romei2014}, a discussion on key choices and assumptions \cite{mitchell2018prediction} and finally \cite{mehrabi2019survey, suresh2019framework} who review different types of biases, \cite{mehrabi2019survey} also introduce a number of methods to mitigate these. In this article we assume that the reader has a working knowledge of applied \ml , i.e. they are familiar with the basic structure of data mining methodologies such as \cite{fayyad1996kdd} and how to apply and evaluate ``standard'' \ml methods. 

Upon this basis, this article aims to: 1) Provide an easy introduction into the area fairness in \ml (\autoref{sec:overview}); 2) Summarise the current approaches to measure fairness in \ml within a standardised notation framework discussing the various trade-offs of each approach as well as their overarching objectives (\autoref{sec:metrics}); 3) Define a two-dimensional taxonomy of approach categories to act as a point of reference. Within this taxonomy, we highlight the main approaches, assumptions, and general challenges for binary classification (\autoref{sec:approaches}) and beyond binary classification (\autoref{sec:beyond}); 4) Highlight currently available toolkits for fair \ml (\autoref{sec:platforms}); and 5) Outline the dilemmas for fairness research as avenues of future work to improve the accessibility of the domain (\autoref{sec:dilemmas}).

\section{Fairness in Machine Learning: key methodological components}
\label{sec:overview}
\simon{should we be defining fairness here? that could be quite a challenge}

\begin{figure}[ht]
    \centering
    \includegraphics[width=.75\textwidth]{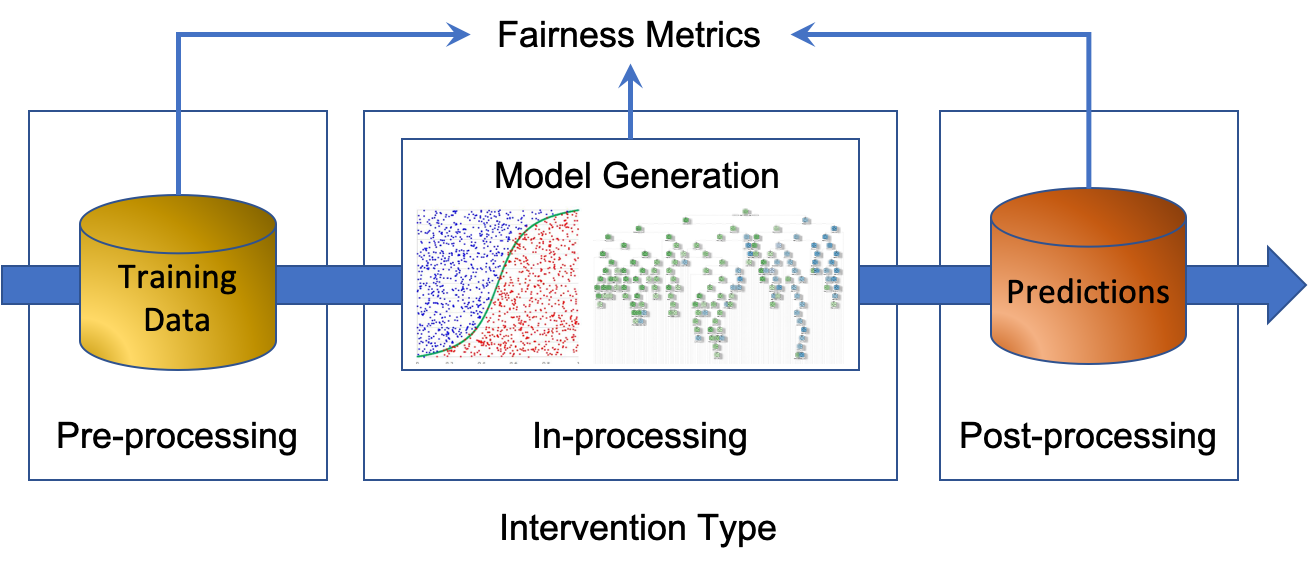}
    \caption{High Level Illustration of Fairness in \ml. Note we omit standard practices common within methodologies like KDD \cite{fayyad1996kdd} to prepare and sample data for \ml methods and their evaluation for visual simplicity.}
    \label{fig:components}
    \tightenfig
\end{figure}

Much of the related literature focuses on either the technical aspects of bias and fairness in \ml, or theorizing on the social, legal, and ethical aspects of \ml discrimination \cite{goodman2016}.  Technical approaches are typically applied prior to modelling (pre-processing), at the point of modelling (in-processing), or after modelling (post-processing), i.e. they emphasize  intervention \cite{binns2017}. In this paper we focus on technical approaches, and in this section give a high-level overview of the framework for an intervention-based methodology for fairness in \ml ; see \autoref{fig:components} for a graphical representation. Whilst not all approaches for fair \ml fit into this framework, it provides a well-understood point of reference and acts as one dimension in a taxonomy of approaches for fairness in \ml .

\subsection{Sensitive and Protected Variables and (Un)privileged Groups} 
\label{sec:variables}
Most approaches to mitigate unfairness, bias, or discrimination are based on the notion of protected or sensitive variables (we will use the terms interchangeably) and on (un)privileged groups: groups (often defined by one or more sensitive variables) that are disproportionately (less) more likely to be positively classified. Before discussing the key components of the fairness framework, a discussion on the nature of protected variables is needed. Protected variables define the aspects of data which are socioculturally precarious for the application of \ml . Common examples are gender, ethnicity, and age (as well as their synonyms). However, the notion of a protected variable can encompass any feature of the data that involves or concerns people \cite{Barocas2019}. 


The question of which variables should be protected quickly arises. We note that many variables are explicitly defined as ``sensitive'' by specific legal frameworks, see \cite{tolan2019machine, lee2018detecting, lepri2018, yeung2018, green2020algorithmic, green2020false, berk2019accuracy, marjanovic2018algorithmic, sokolovska2018integrating} and the references therein. While there are some readily available sets of declared sensitive variables, there are relatively few works that actively seek to determine whether other variables or rare (i.e., minority) combinations should be protected or not. \cite{fisher2018all, adler2018auditing} both present approaches specifically looking at the variable importance (or model influence) of sensitive variables, and could act as a means to identify potentially problematic variables. Yet, there is still the question of variables that are not strictly sensitive, but have a relationship with one or more sensitive variables. \cite{chiappa2018causal} notes that many definitions of fairness express model output in terms of sensitive variables, without considering ``related'' variables. Not considering these related variables could erroneously assume a fair \ml model has been produced. Not considering correlated variables has been shown to increase the risk of discrimination (e.g., redlining\footnote{The term redlining stems from the United States and describes maps that were color coded to represent areas a bank would not invest in, e.g. give loans to residents of these areas \cite{johnson2016impartial}.}) \cite{pedreshi2008, romei2014, zarsky2016, veale2017, dwork2012, calmon2017, du2018data, lum2016, dwork2012, lipton2018mitigating}. 

Understanding ``related'' variables in a general sense is a well studied area, especially in the privacy and data archiving literature, where overlooked variable relationships can enable the deanonymization of published data (see \cite{zimmer2010but}).  The fairness literature, however, often overlooks these effects on fairness, although the relationship between discrimination and privacy was noted in \cite{dwork2012}. In particular, sensitive data disclosure is a long-standing challenge in protecting citizen anonymity when data is published and/or analysed \cite{agrawal2000privacy, fung2010privacy, lindell2000privacy}. Key approaches in this area (e.g. \cite{sweeney2002k, machanavajjhala2007diversity, li2007t}) seek to protect specific individuals and groups from being identifiable within a given dataset, i.e. minimize disclosure risk. Yet, these approaches can still struggle to handle multiple sensitive attributes at once \cite{li2007t}. Whilst these approaches (and many others) have been successful in anonymizing datasets, they still often require a list of features to protect. For explicit identifiers (such as name, gender, zip etc.) such lists exist as already discussed. For correlated or otherwise related variables (often referred to as proxies or quasi-identifiers), much of the literature assumes a priori knowledge of the set of quasi-identifiers \cite{fung2010privacy}, or seeks to discover them on a case-by-case basis (e.g. \cite{motwani2007efficient, jafer2014privacy}), and moves towards a notion of privacy preserving data mining as introduced by \cite{agrawal2000privacy}. \cite{gupta2018proxy} also discuss the notion of proxy groups, a set of ``similar'' instances of the data that could correspond to a protected group (e.g. young women). 

More recently, fairness researchers have begun to investigate graph- and network-based methods for discovering proxies either with respect to anonymity criteria (e.g. \cite{yan2018finding}) or specific notions of fairness (we introduce these approaches in \autoref{sec:repair}). \cite{salimi2019data} provides a brief overview of different theoretical applications to algorithmic fairness, with \cite{glymour2019measuring} noting how different causal graph-based models can help use variable relationships to distill different biases in the model and/or data. \autoref{tbl:proxies} provides some examples of sensitive variables and potential proxies. Ultimately, users need to thoroughly consider how they will identify and define the set of protected variables.



\begin{table}[h]
    \scriptsize
    \centering
    \begin{tabularx}{\textwidth}{p{.165\textwidth}X}
        \toprule
        Sensitive Variable & Example Proxies \\
        \midrule
        Gender &    Education Level, Income, Occupation, Felony Data, Keywords in User Generated Content (e.g. CV, Social Media etc.), University Faculty, Working Hours \\ 
        \midrule
        Marital Status &    Education Level, Income \\
        \midrule
        Race &  Felony Data, Keywords in User Generated Content (e.g. CV, Social Media etc.), Zipcode \\ 
        \midrule
        Disabilities & Personality Test Data \\ 
        \bottomrule        
    \end{tabularx}
    \caption{Example Proxy relationships based on findings from \cite{fisher2018all, berk2019accuracy, salimi2019data, bodie2017law, selbst2017disparate, hall2017say, yarkoni2010personality, schwartz2013toward, weber2014workplace, massey1993american}}
    \label{tbl:proxies}
    \tightentab
\end{table}

\subsection{Metrics} 
Underpinning the intervention-based approaches are an ever increasing array of fairness measures seeking to quantify fairness. The implication of ``measurement'' is, however, precarious as it implies a straightforward process \cite{Barocas2019}. Aside from the philosophical and ethical debates on defining fairness (often overlooked in the \ml literature), creating generalized notions of fairness quantification is challenging. Metrics usually either emphasize individual (e.g. everyone is treated equal), or group fairness, where the latter is further differentiated to within group (e.g. women vs. men) and between group (e.g. young women vs. black men) fairness. Currently, combinations of these ideals using established definitions have been shown to be mathematically intractable \cite{kleinberg2018, chouldechova2017, berk2018fairness}. Quantitative definitions allow fairness to become an additional performance metric in the evaluation of an \ml algorithm. However, increasing fairness often results in lower overall accuracy or related metrics, leading to the necessity of analyzing potentially achievable trade-offs in a given scenario \cite{Haas2019}. 

\subsection{Pre-processing} 
Pre-processing approaches recognize that often an issue is the data itself, and the distributions of specific sensitive or protected variables are biased, discriminatory, and/or imbalanced. Thus, pre-processing approaches tend to alter the sample distributions of protected variables, or more generally perform specific transformations on the data with the aim to remove discrimination from the training data \cite{kamiran2012a}. The main idea here is to train a model on a ``repaired'' data set. Pre-processing is argued as the most flexible part of the data science pipeline, as it makes no assumptions with respect to the choice of subsequently applied modeling technique \cite{du2018data}. 

\subsection{In-processing} 
In-processing approaches recognize that modeling techniques often become biased by dominant features, other distributional effects, or try to find a balance between multiple model objectives, for example having a model which is both accurate and fair. In-processing approaches tackle this by often incorporating one or more fairness metrics into the model optimization functions in a bid to converge towards a model parameterization that maximizes performance and fairness.

\subsection{Post-processing} 
Post-processing approaches recognize that the actual output of an \ml model may be unfair to one or more protected variables and/or subgroup(s) within the protected variable. Thus, post-processing approaches tend to apply transformations to model output to improve prediction fairness. Post-processing is one of the most flexible approaches as it only needs access to the predictions and sensitive attribute information, without requiring access to the actual algorithms and \ml models. This makes them applicable for black-box scenarios where not the entire \ml pipeline is exposed.

\subsection{Initial Considerations: pre-processing vs. in-processing vs. post-processing}
A distinct advantage of pre- and post-processing approaches is that they do not modify the \ml method explicitly. This means that (open source) \ml libraries can be leveraged unchanged for model training. However, they have no direct control over the optimization function of the \ml model itself. Yet, modification of the data and/or model output may have legal implications \cite{barocas2016} and can mean models are less interpretable \cite{lepri2018, lum2016}, which may be at odds with current data protection legislation with respect to explainability. Only in-processing approaches can optimize notions of fairness during model training. Yet, this requires the optimization function to be either accessible, replaceable, and/or modifiable, which may not always be the case.






\section{Measuring Fairness and Bias}
\label{sec:metrics}

Behind intervention-based approaches are a myriad of definitions and metrics (e.g. \cite{barocas2016, berk2018fairness, chouldechova2017, hardt2016, kleinberg2017, woodworth2017, zafar2017}) to mathematically represent bias, fairness, and/or discrimination; but they lack consistency in naming conventions \cite{corbett2018} and notation. More so, there are many different interpretations of what it means for an algorithm to be ``fair''. Several previous publications provide a (limited) overview of multiple fairness metrics and definitions, e.g., \cite{Verma2018, suresh2019framework, pitoura2018measuring, rich2019lessons}. We extend these prior summaries by including additional perspectives for types of biases and a larger set of metrics and definitions that are included as compared to previous publications.





Typically, metrics fall under several categories, for example: 1) \emph{statistical parity}: where each group receives an equal fraction of possible [decision] outcomes \cite{feldman2015, kamishima2012, zemel2013}; 2) \emph{disparate impact}: a quantity that captures whether wildly different outcomes are observed in different [social] groups \cite{feldman2015, zafar2017}; 3) \emph{equality of opportunity} \cite{hardt2016}, 4) \emph{calibration} \cite{kleinberg2017}: where false positive rates across groups are enforced to be similar (defined as \emph{disparate mistreatment} by \cite{zafar2017} when this is not the case), 5) \emph{counterfactual fairness} which states that a decision is fair towards an individual if it coincides with one that would have been taken were the sensitive variable(s) different \cite{chiappa2019path}. 

Although the literature has defined a myriad of notions to quantify fairness, each measures and emphasizes different aspects of what can be considered ``fair''. Many are difficult / impossible to combine \cite{kleinberg2018, chouldechova2017}, but ultimately, we must keep in mind (as noted in \cite{chierichetti2019matroids}) there is no universal means to measure fairness, and also at present no clear guideline(s) on which measures are ``best''. Thus, in this section we provide an overview of fairness measures and seek to provide a lay interpretation to help inform decision making. Table \ref{tbl:metricoverview} presents an overview of the categories of fairness metrics presented with \autoref{tbl:notationclassification} introducing key notation. 

\begin{table}[h]
    \scriptsize
    \centering
    \begin{tabularx}{\textwidth}{p{.07\textwidth}@{\hskip 0.15in}X@{\hskip 0.15in}X@{\hskip 0.15in}X@{\hskip 0.15in}X@{\hskip 0.15in}p{0.15\textwidth}}
        \toprule
        & \multicolumn{4}{c}{Group-based Fairness} & Individual and Counterfactual Fairness \\
        \cmidrule(lr){2-5} \cmidrule(lr){6-6}
         & Parity-based Metrics & Confusion Matrix-based Metrics & Calibration-based Metrics & Score-based Metrics & Distribution-based Metrics  \\
        \midrule
        Concept & Compare predicted positive rates across groups & Compare groups by taking into account potential underlying differences between groups & Compare based on predicted probability rates (scores) & Compare based on expected scores & Calculate distributions based on individual classification outcomes \\
        Abstract Criterion & Independence & Separation & Sufficiency & - & - \\
        Examples & Statistical Parity, Disparate Impact & Accuracy equality, Equalized Odds, Equal Opportunity & Test fairness, Well calibration & Balance for positive and negative class, Bayesian Fairness & Counterfactual Fairness, Generalized Entropy Index \\
        \bottomrule        
    \end{tabularx}
    \caption{Overview of suggested fairness metrics for binary classification}
    \label{tbl:metricoverview}
    \tightentab
\end{table}


\newcommand\MyBox[2]{
  \fbox{\lower0.75cm
    \vbox to 1.7cm{\vfil
      \hbox to 1.7cm{\hfil\parbox{1.4cm}{#1\\#2}\hfil}
      \vfil}%
  }%
}


\begin{table}{}
    \centering
    \scriptsize
    \begin{tabularx}{0.8\textwidth}{p{3cm}X}  
        \toprule
        Symbol   & Description \\
        \midrule
        $y\in {0, 1}$ & Actual value / outcome \\
        $\hat{y}\in {0, 1}$ & Predicted value / outcome \\
        $s=Pr(\hat{y}_i=1)$ & Predicted score of an observation $i$. Probability of $y=1$ for observation $i$ \\
        $g_i, g_j$ & Identifier for groups based on protected attribute \\
        \bottomrule
    \end{tabularx}
    \caption{Notation for Binary Classification}
    \label{tbl:notationclassification}
    \tightentab
\end{table}

\subsection{Abstract Fairness Criteria}
\label{subsec:abstractfairnesscriteria}
Most quantitative definitions and measures of fairness are centered around three fundamental aspects of a (binary) classifier: First, the sensitive variable $S$ that defines the groups for which we want to measure fairness. Second, the target variable $Y$. In binary classification, this represents the two classes that we can predict: $Y=0$ or $Y=1$. Third, the classification score $R$, which represents the predicted score (within $[0,1]$) that a classifier yields for each observation. Using these properties, general fairness desiderata are categorized into three ``non-discrimination'' criteria \cite{Barocas2019}:


\textbf{Independence} aims for classifiers to make their scoring independent of the group membership:
\begin{align}
    R \perp S
\end{align}

An example group fairness metric focusing on independence is Statistical/Demographic Parity. Independence does not take into account that the outcome $Y$ might be correlated with the sensitive variable $S$. I.e., if the separate groups have different underlying distributions for $Y$, not taking these dependencies into account can lead to outcomes that are considered fair under the Independence criterion, but not for (some of the) groups themselves. Hence, an extension of the Independence property is the \textbf{Separation} criterion which looks at the independence of the score and the sensitive variable conditional on the value of the target variable $Y$:
\begin{align}
    R \perp S | Y
\end{align}
Example metrics that target the Separation property are Equalized Odds and Equal Opportunity. The third criterion commonly used is \textbf{Sufficiency}, which looks at the independence of the target $Y$ and the sensitive variable $S$, conditional for a given score $R$:
\begin{align}
    Y \perp S | R
\end{align}

As \cite{Barocas2019} point out, Sufficiency is closely related to some of the calibration-based metrics. 
\cite{Barocas2019} also discuss several impossibility results with respect to these three criteria. For example, they show that if $S$ and $Y$ are not independent, then Independence and Sufficiency cannot both be true. This falls into a more general discussion on impossibility results between fairness metrics.
\christian{Should we extend this? I didn't have a chance to look at this particular aspect, but I think at least mentioning that people work on this might be worthwhile. Also, I'm mainly citing the FairML book here which provides great descriptions, but I'm not 100\% sure if they are the original definitions.}

\subsection{Group Fairness Metrics}

Group-based fairness metrics essentially compare the outcome of the classification algorithm for two or more groups. Commonly, these groups are defined through the sensitive variable as described in \autoref{sec:variables}. Over time, many different approaches have been suggested, most of which use metrics based on the binary classification confusion matrix to define fairness.

\subsubsection{Parity-based Metrics}\hfill\\
Parity-based metrics typically consider the predicted positive rates, i.e., $Pr(\hat{y}=1)$, across different groups. This is related to the Independence criterion that was defined in \autoref{subsec:abstractfairnesscriteria}.

\textbf{Statistical/Demographic Parity}: 
One of the earliest definitions of fairness, this metric defines fairness as an equal probability of being classified with the positive label \cite{zemel2013,kamishima2012,feldman2015,corbett2017}. I.e., each group has the same probability of being classified with the positive outcome. A disadvantage of this notion, however, is that potential differences between groups are not being taken into account. 
\begin{align}
    Pr(\hat{y}=1 | g_i) = Pr(\hat{y}=1 | g_j)
\end{align}

\textbf{Disparate Impact}: Similar to statistical parity, this definition looks at the probability of being classified with the positive label. In contrast to parity, Disparate Impact considers the ratio between unprivileged and privileged groups. Its origins are in legal fairness considerations for selection procedures which sometimes use an 80\% rule to define if a process has disparate impact (ratio smaller than 0.8) or not \cite{feldman2015}. 
\begin{align}
   \frac{Pr(\hat{y}=1 | g_1)}{Pr(\hat{y}=1 | g_2)} 
\end{align}
While often used in the (binary) classification setting, notions of Disparate Impact are also used to define fairness in other domains, e.g., dividing a finite supply of items among participants \cite{peysakhovich2019fair}.

\subsubsection{Confusion Matrix-based Metrics}\hfill\\
While parity-based metrics typically consider variants of the predicted positive rate $Pr(\hat{y}=1)$, confusion matrix-based metrics take into consideration additional aspects such as \tpr, \tnr, \fpr, and \fnr. The advantage of these types of metrics is that they are able to include underlying differences between groups who would otherwise not be included in the parity-based approaches. This is related to the Separation criterion that was defined in \autoref{subsec:abstractfairnesscriteria}.

\textbf{Equal Opportunity}: As parity and disparate impact do not consider potential differences in groups that are being compared, \cite{hardt2016,pleiss2017} consider additional metrics that make use of the \fpr and \tpr between groups. Specifically, an algorithm is considered to be fair under equal opportunity if its \tpr is the same across different groups. 
\begin{align}
   Pr(\hat{y}=1 |y=1 \& g_i) &= Pr(\hat{y}=1 | y=1 \& g_j) 
\end{align}

\textbf{Equalized Odds} (Conditional procedure accuracy equality \cite{berk2018fairness}): Similarly to equal opportunity, in addition to \tpr equalized odds simultaneously considers \fpr as well, i.e., the percentage of actual negatives that are predicted as positive. 
\begin{align}
   Pr(\hat{y}=1 |y=1 \& g_i) = Pr(\hat{y}=1 | y=1 \& g_j) &\And Pr(\hat{y}=1 |y=0 \& g_i) = Pr(\hat{y}=1 | y=0 \& g_j)
\end{align}


\textbf{Overall accuracy equality} \cite{berk2018fairness}: Accuracy, i.e., the percentage of overall correct predictions (either positive or negative), is one of the most widely used classification metrics. \cite{berk2018fairness} adjusts this concept by looking at relative accuracy rates across different groups. If two groups have the same accuracy, they are considered equal based on their accuracy. 
\begin{align}
   Pr(\hat{y}=0 |y=0 \& g_i) + Pr(\hat{y}=1 |y=1 \& g_i) &= Pr(\hat{y}=0 |y=0 \& g_j) + Pr(\hat{y}=1 |y=1 \& g_j)
\end{align}

\textbf{Conditional use accuracy equality} \cite{berk2018fairness}: As an adaptation of the overall accuracy equality, the following conditional procedure and conditional use accuracy do not look at the overall accuracy for each subgroup, but rather at the positive and negative predictive values. 
\begin{align}
   Pr(y=1 | \hat{y}=1 \& g_i) &= Pr(y=1 | \hat{y}=1 \& g_j) \And 
   Pr(y=0 | \hat{y}=0 \& g_i) &= Pr(y=0 | \hat{y}=0 \& g_j)
\end{align}

\textbf{Treatment equality} \cite{berk2018fairness}: Treatment equality considers the ratio of False Negative Predictions (\fnr) to False Positive Predictions. 
\begin{align}
  \frac{Pr(\hat{y}=1 |y=0 \& g_i)}{Pr(\hat{y}=0 |y=1 \& g_i)} &= \frac{Pr(\hat{y}=1 |y=0 \& g_j)}{Pr(\hat{y}=0 |y=1 \& g_j)}
\end{align}

\textbf{Equalizing disincentives} \cite{jung2020fair}: The Equalizing disincentives metric compares the difference of two metrics, \tpr and \fpr, across the groups and is specified as:
\begin{align}
   Pr(\hat{y}=1 |y=1 \& g_i) - Pr(\hat{y}=1 |y=0 \& g_i) &= Pr(\hat{y}=1 |y=1 \& g_j) - Pr(\hat{y}=1 |y=0 \& g_j)
\end{align}

\textbf{Conditional Equal Opportunity} \cite{beutel2019putting}: As some metrics can be dependent on the underlying data distribution, \cite{beutel2019putting} provide an additional metric that specifies equal opportunity on a specific attribute $a$ our of a list of attributes $A$, where $\tau$ is a threshold value:
\begin{align}
   Pr(\hat{y}\geq \tau |g_i \& y<\tau \& A=a) &= Pr(\hat{y}= \geq \tau | g_j \& y<\tau \& A=a)
\end{align}

\subsubsection{Calibration-based Metrics}\hfill\\
In comparison to the previous metrics which are defined based on the predicted and actual values, 
calibration-based metrics take the predicted probability, or score, into account. This is related to the Sufficiency criterion that was defined in Section \ref{subsec:abstractfairnesscriteria}.

\textbf{Test fairness/ calibration / matching conditional frequencies} (\cite{chouldechova2017}, \cite{hardt2016}): Essentially, test fairness or calibration wants to guarantee that the probability of $y=1$ is the same given a particular score. I.e., when two people from different groups get the same predicted score, they should have the same probability of belonging to $y=1$. 
\begin{align}
   Pr(y=1 | S=s \& g_i) &= Pr(y=1 | S=s \& g_j) 
\end{align}

\textbf{Well calibration} \cite{kleinberg2017}: An extension of regular calibration where the probability for being in the positive class also has to equal the particular score.
\begin{align}
   Pr(y=1 | S=s \& g_i) &= Pr(y=1 | S=s \& g_j) = s
\end{align}

\subsubsection{Score-based Metrics}\hfill\\
\textbf{Balance for positive and negative class} \cite{kleinberg2017}: The expected predicted score for the positive and negative class has to be equal for all groups:
\begin{align}
   E(S=s | y=1 \& g_i) &= E(S=s | y=1 \& g_j), E(S=s | y=0 \& g_i) = E(S=s | y=0 \& g_j)
\end{align}

\textbf{Bayesian Fairness} \cite{dimitrakakis2019bayesian} extend the balance concept from \cite{kleinberg2017} when model parameters themselves are uncertain. Bayesian fairness considers scenarios where the expected utility of a decision maker has to be balanced with fairness of the decision. The model takes into account the probability of different scenarios (model parameter probabilities) and the resulting fairness / unfairness.  

\subsection{Individual and Counterfactual Fairness Metrics}
As compared to group-based metrics which compare scores across different groups, individual and counterfactual fairness metrics do not focus on comparing two or more groups as defined by a sensitive variable, but consider the outcome for each participating individual. \cite{Kusner2017} propose the concept of counterfactual fairness which builds on causal fairness models and is related to both individual and group fairness concepts. \cite{speicher2018} proposes a generalized entropy index which can be parameterized for different values of $\alpha$ and measures the individual impact of the classification outcome. This is similar to established distribution indices such as the Gini Index in economics.

\textbf{Counterfactual Fairness}: Given a causal model $(U,V,F)$, where $U$ are latent (background) variables, $V = S \cup X$ are observable variables including the sensitive variable $S$, and $F$ is a set of functions defining structural equations such that $V$ is a function of $U$, counterfactual fairness is:
\begin{align} 
    P(\hat{y}_{A \leftarrow a}(U) = y | X = x, A=a ) = P(\hat{y}_{A \leftarrow a^{'}}(U) = y | X = x, A=a )
\end{align}
Essentially, the definition ensures that the prediction for an individual coincides with the decision if the sensitive variable would have been different.

\textbf{Generalized Entropy Index}: \cite{speicher2018} defines the Generalized Entropy Index (GEI) which considers differences in an individual's prediction ($b_i$) to the average prediction accuracy ($\mu$). It can be adjusted based on the parameter $\alpha$, where $b_i=\hat{y}_i-y_i+1$ and $\mu=\frac{\sum_i b_i}{n}$: 
\begin{align}
    GEI = \frac{1}{n\alpha(\alpha-1)}\sum_{i=1}^{n}\left[ (\frac{b_i}{\mu})^\alpha -1\right]
\end{align}

\textbf{Theil Index}: a special case of the GEI for $\alpha = 1$. In this case, the calculation simplifies to:
\begin{align}
    Theil = \frac{1}{n}\sum_{i=1}^{n}(\frac{b_i}{\mu})log(\frac{b_i}{\mu}))
\end{align}

\subsection{Summary}
The literature is at odds with respect to whether individual or group fairness should be prioritized. \cite{speicher2018} note that many approaches to group fairness tackle only between-group issues, as a consequence they demonstrate that within-group issues are worsened through this choice. Consequently, users must decide on where to place emphasis, but be mindful of the trade off between any fairness measure and model accuracy \cite{berk2017, dwork2012, corbett2017, hardt2016, zliobaite2015, calmon2017}. With a reliance on expressing fairness and bias mathematically, \cite{corbett2018, green2018} argue that these definitions often do not map to normative social, economic, or legal understandings of the same concepts. This is corroborated by \cite{skirpan2017} who note an over emphasis in the literature on disparate treatment. \cite{agarwal2018, calmon2017, speicher2018} criticize ad hoc and implicit choices concerning distributional assumptions or realities of relative group sizes.
\section{Binary Classification Approaches}
\label{sec:approaches}



Building on the metrics discussed in \autoref{sec:metrics}, fairness in \ml researchers seek to mitigate unfairness by ``protecting'' sensitive sociodemographic attributes (as introduced in \autoref{sec:variables}). The literature is dominated by approaches for mitigating bias and unfairness in \ml within the problem class of binary classification \cite{berk2017}. There are many reasons for this, but most notably: 1) many of the most contentious application areas that motivated the domain are binary decisions (hiring vs. not hiring; offering a loan vs. not offering a loan; re-offending vs. not re-offending etc.); 2) quantifying fairness on a binary dependent variable is mathematically more convenient; addressing multi-class problems would at the very least add terms in the fairness quantity. 

\afterpage{
\begin{landscape}
\begin{center}
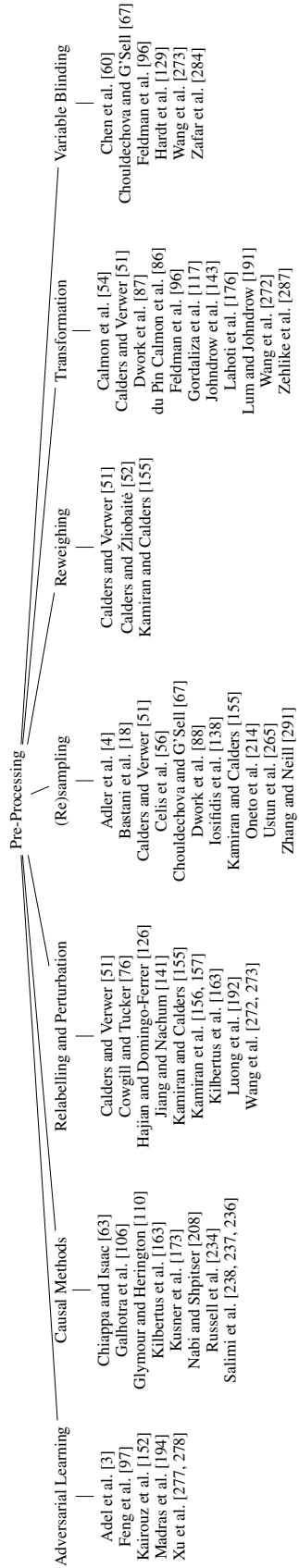
\begin{figure}[htp]
    \begin{adjustbox}{width=\linewidth}
    \begin{forest}
        [Pre-Processing
            [Adversarial Learning [{\begin{tabular}{c}
                \citet{adel2019one}\\
                \citet{feng2019learning}\\
                \citet{kairouz2019censored}\\
                \citet{madras2018}\\
                \citet{xu2018fairgan, xu2019achieving}
            \end{tabular}}]]        
            [Causal Methods [{\begin{tabular}{c}        
                \citet{chiappa2018causal}\\
                \citet{galhotra2017fairness}\\
                \citet{glymour2019measuring}\\
                \citet{kilbertus2019fair}\\
                \citet{Kusner2017}\\
                \citet{nabi2018fair}\\
                \citet{russell2017worlds}\\
                \citet{salimi2019interventional, salimi2019capuchin, salimi2019data}
            \end{tabular}}]]
            [Relabelling and Perturbation [{\begin{tabular}{c}
                \citet{calders2010}\\
                \citet{cowgill2017}\\
                \citet{hajian2012methodology}\\
                \citet{jiang2019identifying}\\
                \citet{kamiran2012a}\\
                \citet{kamiran2010, kamiran2012b}\\
                \citet{kilbertus2019fair}\\
                \citet{luong2011} \\
                \citet{wang2018avoiding, wang2019repairing}\\
            \end{tabular}}]]
            [(Re)sampling [{\begin{tabular}{c}        
                \citet{adler2018auditing}\\
                \citet{bastani2019probabilistic}\\
                \citet{calders2010}\\
                \citet{celis2016}\\
                \citet{chouldechova2017fairer}\\
                \citet{dwork2018}\\
                \citet{iosifidis2020fae}\\
                \citet{kamiran2012a}\\
                \citet{oneto2019taking}\\
                \citet{ustun2019fairness}\\
                \citet{zhang2016identifying}\\
            \end{tabular}}]]
            [Reweighing [{\begin{tabular}{c}
                \citet{calders2010}\\
                \citet{Calders2013}\\
                \citet{kamiran2012a}\\
            \end{tabular}}]]
            [Transformation [{\begin{tabular}{c}
                \citet{calmon2017}\\
                \citet{calders2010}\\
                \citet{dwork2012}\\
                \citet{du2018data}\\
                \citet{feldman2015}\\
                \citet{gordaliza2019obtaining}\\
                \citet{johndrow2019algorithm}\\
                \citet{lahoti2019ifair}\\
                \citet{lum2016}\\
                \citet{wang2018avoiding}\\
                \citet{zehlike2019continuous}\\
            \end{tabular}}]]     
            [Variable Blinding [{\begin{tabular}{c}
                \citet{chen2018discriminatory}\\
                \citet{chouldechova2017fairer}\\
                \citet{feldman2015}\\
                \citet{hardt2016}\\
                \citet{wang2019repairing}\\
                \citet{zafar2017}\\
            \end{tabular}}]]
        ] 
    \end{forest}
    \end{adjustbox}
    \vspace{-0.5cm}
    \caption{Pre-processing Methods}
    \label{fig:pre}
\end{figure}

\vspace*{-0.55cm}

\begin{figure}[htp]
\begin{adjustbox}{width=\linewidth}
\begin{forest}
        [In-Processing
            [Adversarial Learning [{\begin{tabular}{c}
                \citet{beutel2019putting, beutel2017}\\
                \citet{celis2019improved}\\
                \citet{edwards2015}\\
                \citet{feng2019learning}\\
                \citet{wadsworth2018achieving}\\ 
                \citet{xu2019achieving}\\
                \citet{zhang2018}\\
            \end{tabular}}]]      
            [Bandits [{\begin{tabular}{c}
                \citet{ensign2018decision}\\
                \citet{gillen2018online}\\
                \citet{joseph2016fairness, joseph2018meritocratic}\\
                \citet{liu2017calibrated}\\
            \end{tabular}}]]
            [Constraint Optimization [{\begin{tabular}{c c}
                \citet{Celis2019} & 
                \citet{chierichetti2019matroids}\\
                \citet{cotter2019optimization} &
                \citet{goh2016satisfying}\\
                \citet{Haas2019} &
                \citet{kim2018fairness}\\
                \citet{manisha2018neural} &
                \citet{nabi2018fair}\\
                \citet{nabi2019learning} &
                \citet{narasimhan2018learning} \\
                \citet{zemel2013} & \\
            \end{tabular}}]]
            [Regularization [{\begin{tabular}{c c}
                \citet{aghaei2019learning}&
                \citet{bechavod2017penalizing}\\
                \citet{berk2017}&
                \citet{di2020counterfactual}\\
                \citet{feldman2015}&
                \citet{goel2018non}\\
                \citet{heidari2018fairness}&
                \citet{huang2019stable}\\
                \citet{jiang2019wasserstein} & 
                \citet{kamishima2012}\\
            \end{tabular}}]]
            [Reweighing [{\begin{tabular}{c}
                \citet{krasanakis2018adaptive}\\
                \citet{jiang2019identifying}\\
            \end{tabular}}]]
        ] 
    \end{forest}
    \end{adjustbox}
    \vspace{-.5cm}
    \caption{In-processing Methods}
    \label{fig:inproc}
\end{figure}
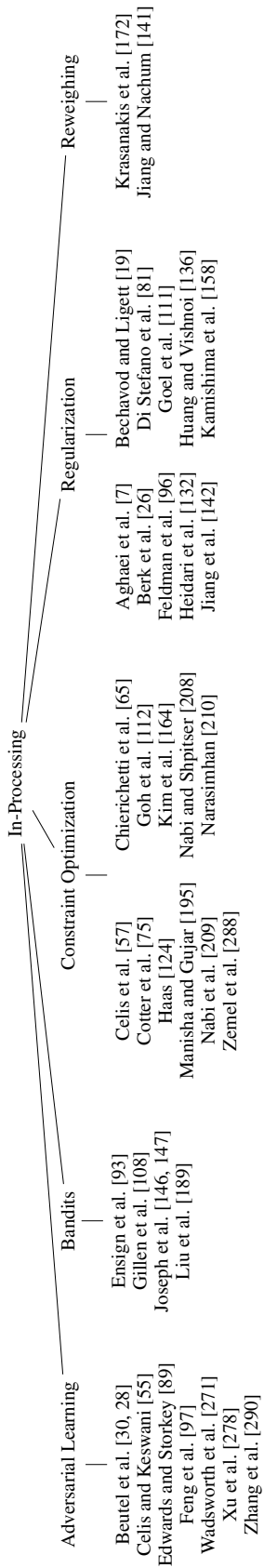
\end{center}

\vspace*{-.5cm}

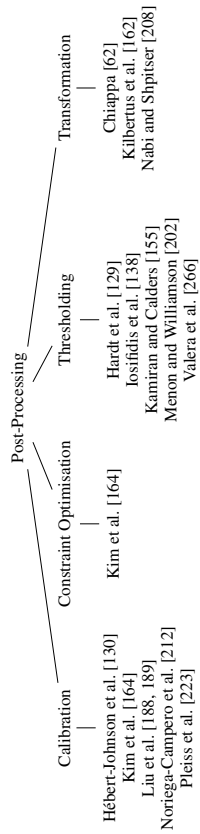
\begin{figure}[htp]
\small
    \begin{adjustbox}{width=0.5\linewidth}
    \begin{forest}
        [Post-Processing
            [Calibration [{\begin{tabular}{c}
                \citet{hebert2017calibration}\\
                \citet{kim2018fairness}\\
                \citet{liu2019implicit, liu2017calibrated}\\
                \citet{noriega2019active}\\
                \citet{pleiss2017}\\
            \end{tabular}}]]
            [Constraint Optimisation [{\begin{tabular}{c}
                \citet{kim2018fairness} \\
            \end{tabular}}]]
            [Thresholding [{\begin{tabular}{c}
                \citet{hardt2016}\\
                \citet{iosifidis2020fae}\\
                \citet{kamiran2012a}\\
                \citet{menon2017cost}\\
                \citet{valera2019enhancing}\\
            \end{tabular}}]]
            [Transformation [{\begin{tabular}{c}
                \citet{chiappa2019path}\\
                \citet{Kilbertus2017}\\
                \citet{nabi2018fair}\\
            \end{tabular}}]]
        ] 
    \end{forest}
    \end{adjustbox}
    \caption{Post-processing methods}
    \label{fig:postproc}
\end{figure}

\end{landscape}}

In this section, we discuss the main approaches for tackling fairness in the binary classification case. We begin by arranging mitigation methods into a visual taxonomy according to the location in the \ml framework (\autoref{fig:components}), i.e. pre-processing: \autoref{fig:pre}, in-processing: \autoref{fig:inproc}, and post-processing: \autoref{fig:postproc}. We note an abundance of pre- and in-processing vs. post-processing methods and that method families, e.g. methods leveraging adversarial learning, can belong to multiple stages (pre- and in-processing in this case). It is also noteworthy that many of the approaches listed (i.e. the overall mitigation strategy applied by researchers) do not belong to a single category or stage, but several: approaches tend to be hybrid and this is becoming more prevalent in more recent approaches. However, we are yet to find an approach using methods from all three stages, even if there are papers comparing methods from multiple stages. Finally, we also note that we do not comment on the advantages of specific approaches over others, yet where relevant we outline challenges researchers must navigate. The literature has an urgent need for a structured meta review of approaches to fairness. Whilst many papers compare specific subsets of the approaches in this section, they do not and realistically cannot offer a holistic comparison.

\subsection{Blinding}
\label{sec:blinding}
Blinding is the approach of making a classifier ``immune'' to one or more sensitive variables \cite{zeng2017interpretable}. A classifier is, for example, race blind if there is no observable outcome differentiation based on the variable race. For example, \cite{hardt2016} seek to train a race blind classifier (among others) in that each of the 4 race groups have the same threshold value (see \autoref{sec:thresholding}), i.e. the provided loan rate is equal for all races. Other works have termed the omission of sensitive variables from the training data as blinding. However, we distinguish {\bf immunity} to sensitive variables as distinct from {\bf omission of} sensitive variables. Omission has been shown to decrease model accuracy \cite{hajian2012, chen2018discriminatory, salimi2019interventional} and not improve model discrimination \cite{calmon2017, kamiran2012a, dwork2012}. Both omission and immunity overlook relationships with proxy variables (as discussed in \autoref{sec:variables}, we note \cite{wang2019repairing} as an exception here who omit proxies), which can result in increasing instead of decreasing bias and discrimination \cite{kleinberg2018}, or indirectly concealing discrimination \cite{du2018data}. It also ignores that discrimination may not be one variable in isolation, but rather the result of several joint characteristics \cite{pedreshi2008} and as such determining which combination(s) of variables to blind is non-trivial and the phenomenon of omitted variable bias should not be downplayed \cite{clarke2005phantom, jung2018omitted}.

Approaches in subgroup analysis (see \autoref{sec:sampling}) have used statistical techniques to determine when variable immunity does not adversely affect fairness (e.g. \cite{chouldechova2017fairer}). Similarly, researchers still use omission in their evaluation methodologies to compare to earlier works and act as a baseline. Omission can also be used in specific parts of the fairness methodology, for example \cite{feldman2015} temporarily omit sensitive variables prior to transforming (see \autoref{sec:transformation}) the training data. Blinding (or partial blinding) has also been used as a fairness audit mechanism \cite{adler2018auditing, henelius2014peek, datta2016algorithmic}. Specifically, such approaches explore how partially blinding features (sensitive or otherwise) affect model performance. This is similar to the idea of causal models (\autoref{sec:repair}) and can help identify problematic sensitive or proxy variables with black-box-like analysis of a \ml model.
\subsection{Causal Methods}
\label{sec:repair}
Approaches using causal methods recognise that the data upon which \ml models are trained often reflect some form of underlying discrimination. A key objective is to uncover causal relationships in the data and find dependencies between sensitive and non-sensitive variables \cite{galhotra2017fairness, kilbertus2019fair, Kusner2017, chiappa2018causal, glymour2019measuring, nabi2018fair, russell2017worlds}. Thus, causal methods are specifically well suited to identifying proxies of sensitive variables as discussed in \autoref{sec:variables} as well as subgroup analyses of which subgroups are most (un)fairly treated and differentiate the types of bias exhibited \cite{glymour2019measuring}. Alternatively, casual methods can be employed to provide transparency with respect to how (classification) decisions were made \cite{henelius2014peek}. In both scenarios, they provide visual descriptions of (un)fairness in the dataset upon the basis of bias in terms of causation (see \cite{Kilbertus2017, Kusner2017, kusner2018causal, henelius2014peek}). Directed acyclic graphs (DAGs) are a common means to represent conditional independence assumptions between variables \cite{johnson2016impartial}. 

Extensions have also leveraged causal dependencies to ``repair'' training data \cite{salimi2019interventional, salimi2019capuchin, salimi2019data} using dependency information to repair (insert, modify, and remove) samples from the training data in accordance to satisfying fairness-specific constraints and conditional independence properties of the training data. Initial results with data repair methods have shown to result in ``debiased'' classifiers which are robust to unseen test data, yet require significant computational resources. The main challenge for causal models is that they require background information and context regarding the causal model which may not always be accessible \cite{salimi2019interventional}. They have also been criticized for not well examining how they would be applied in practice \cite{salimi2019interventional}. 
\subsection{Sampling and Subgroup Analysis}
\label{sec:sampling}

Sampling methods have two primary objectives: 1) to create samples for the training of robust algorithms (e.g. \cite{celis2016, zafar2017, dwork2018, ustun2019fairness, alabi2018unleashing, iosifidis2020fae}), i.e. seek to ``correct'' training data and eliminate biases \cite{iosifidis2020fae}; and 2) to identify groups (or subsamples) of the data that are significantly disadvantaged by a classifier, i.e. as a means to evaluate a model (e.g. \cite{chouldechova2017fairer, adler2018auditing, zhang2016identifying}) via subgroup analysis. \cite{celis2016} articulate the challenge of sampling for fairness as the following question: how are samples selected from a (large) dataset that are both diverse in features and fair to sensitive attributes? Without care, sampling can propagate biases within the training data \cite{oneil2016, dwork2018}, as ensuring diversity in the data used to train the model makes no guarantees of producing fair models \cite{oneil2016}. As such, approaches that seek to create fair training samples include notions of fairness in the sampling strategy. \cite{kamiran2012a}, propose to preferentially sample (similar to oversampling) instances ``close'' to a decision boundary (based on an initial model prototype to approximate a decision boundary) as these are most likely to be discriminated or favored due to underlying biases within the training data.

Within the sampling approaches, the application of decoupled classifiers and multitask learning has emerged (see \cite{zafar2017, dwork2018, ustun2019fairness, alabi2018unleashing, oneto2019taking, calders2010}). Here, the training data is split into subgroups (decoupled classifiers), i.e. combinations of one or more sensitive variables (e.g. [old, males]), or these groupings are learned in a preprocessing stage (multitask learning). Thus, such approaches seek to make the most accurate models for given subgroups (decoupled classifiers) or considering the observation of different subgroups (multitask learning).\footnote{Whilst not a sampling method, learning proxy groupings \cite{gupta2018proxy} is similar in intent to applications of multitask learning.} \cite{iosifidis2020fae} have taken this approach a little further and create an ensemble of ensembles where each base level ensemble operates on a given protected group. Note that sufficient data is required for each subgroup for this method of sampling to not negatively affect performance and fairness, as shown by \cite{alabi2018unleashing}, and that outliers can be problematic \cite{brooks2011support, nguyen2013algorithms}. For each subgroup, an individual classifier is trained. 

A key challenge in decoupled classifiers is the selection of groups: some can be rarer than others \cite{ustun2019fairness} and as such a balance is needed to ensure groups are as atomic as possible but robust against gerrymandering \cite{hebert2017calibration, kearns2018preventing}. Thus, different candidate groupings are often evaluated via in- or post-processing methods to inhibit overfitting, maximize some fairness metric(s), and/or prevent other theoretical violations. Common approaches in group formation are recursive partitioning (e.g. \cite{woodworth2017, kearns2018preventing}) and clustering (e.g.: \cite{chouldechova2017fairer, iosifidis2020fae}) as (good) clusters well approximate underlying data distributions and subgroups. \cite{iosifidis2020fae} used clustering as a means to build stratified samples of different subgroups within the data as an exercise in bagging for the training of fair ensembles. 

Subgroup analysis can also be a useful exercise in model evaluation \cite{chouldechova2017fairer, adler2018auditing, zhang2016identifying} often defining quantities to measure how models affect different subgroups. For example, to analyze if one model is more discriminatory than another to some observed subgroup, or to identify how variable omission (see \autoref{sec:blinding}) affects model fairness. Statistical hypothesis testing is employed to reveal whether models are significantly different with respect to fairness quantities or denote variable instability, i.e. when a model is not robustly fair when a given variable or set of variables are included. These methods can also treat previously trained \ml models as a black-box \cite{chouldechova2017fairer, adler2018auditing}. See \cite{mcnair2018preventing} for an example set of statistical tests to indicate the likelihood of fair decisions. Probabilistic verification (e.g. \cite{bastani2019probabilistic}) of fairness measures via the sampling of sensitive variables has also been proposed to evaluate a trained model within some (small) confidence bound. Similar to other evaluation approaches (e.g. \cite{albarghouthi2017fairsquare}), these approaches present fairness as a dichotomous outcome: a model is fair, or it isn't, as opposed to quantifying how (un)fair a model is. Yet, this is still a useful (and scalable) means to quickly evaluate different models against a number of fairness metrics.
\subsection{Transformation}
\label{sec:transformation}
Transformation approaches learn a new representation of the data, often as a mapping or projection function, in which fairness is ensured, but still preserving the fidelity of the \ml task \cite{feldman2015}. Current transformation approaches operate mainly on numeric data, which is a significant limitation \cite{feldman2015}. There are different perspectives to transforming the training data: operating on the dependent variable (e.g. \cite{du2018data}), operating on numeric non-sensitive variables (e.g. \cite{feldman2015, lum2016, calders2010}), mapping individuals to an input space which is independent of specific protected subgroupings (e.g. \cite{dwork2012, zemel2013, calmon2017, lahoti2019ifair, gordaliza2019obtaining, johndrow2019algorithm, zehlike2019continuous}), and transforming the distribution of model predictions in accordance to specific fairness objectives (e.g. \cite{jiang2019wasserstein, wang2018avoiding}). There are parallels between blinding (in the immunity sense) and independence mappings, as in many ways these two approaches share a common goal: creating independence from one or more specific sensitive variables. Other forms of transformation include relabelling and perturbation, but we consider these a class of their own (see: \autoref{sec:labelling}). 

To provide an example for transformation, \cite{feldman2015} discuss transforming the distribution of SAT scores towards the median to ``degender'' the original distribution into a distribution which retains only the rank order for individuals independent of gender. This is essentially removing information about protected variables from a set of covariates. 
Transformation approaches often seek to retain rank orders within transformed variables in order to preserve predictive ability. 
\cite{calders2010, lum2016} define a similar approach yet model the transformation process with different assumptions and objectives. An alternative to retaining rank order is the use of distortion constraints (e.g. \cite{du2018data}) which seek to prevent mapping ``high'' values to ``low'' values and vice versa. There is an inherent trade-off between the degree of transformation (fairness repair) and the effect on classifier performance \cite{feldman2015}. Approaches to combat this often resort to partial repair: transforming the data towards some target distribution, but not in its entirety seeking to balance this trade-off (e.g. \cite{feldman2015, gordaliza2019obtaining}).

Although largely a pre-processing method, transformation can also been applied within a post-processing phase. \cite{chiappa2019path, Kilbertus2017, nabi2018fair} transform the output of a classifier in accordance to the identification of unfair causal pathways, either by averaging \cite{nabi2018fair}, constraining the conditional distribution of the decision variable \cite{Kilbertus2017} or through counterfactual correction \cite{chiappa2019path}. As an approach, this is similar to the idea of thresholding (see \autoref{sec:thresholding}) and calibration (see \autoref{sec:calibration}). 

There are a number of challenges to consider when applying transformation techniques: 1) the transformed data should not be significantly different from the original data, otherwise the extent of ``repair'' can diminish the utility of the produced classifier \cite{lum2016, feldman2015, du2018data} and, in general, incur data loss \cite{gordaliza2019obtaining}; 2) understanding the relationships between sensitive and potential proxy variables \cite{feldman2015}, as such \ml researchers may wish to use causal methods to first understand these relationships prior to the application of transformation methods; 3) the selection of ``fair'' target distributions is not straightforward \cite{du2018data, zliobaite2015, gordaliza2019obtaining}; 4) finding an ``optimal'' transformation often requires an optimisation step, which under high dimensionality can be computationally expensive, even under assumptions of convexity \cite{du2018data}; 5) missing data provides specific problems for transformation approaches, as it is unclear how to deal with such data samples. Many handle this by simply removing these samples, yet this may raise other methodological issues; 6) transformation makes the model less interpretable \cite{lepri2018, lum2016}, which may be at odds with current data protection legislation; and 7) there are no guarantees that a transformed data set has ``repaired'' potentially discriminatory latent relationships with proxy variables \cite{calmon2017}. In this case, causal methods (see \autoref{sec:repair}) may help.
\subsection{Relabelling and Perturbation}
\label{sec:labelling}

Relabelling and perturbation approaches are a specific subset of transformation approaches: they either flip or modify the dependent variable (relabelling; e.g.\cite{cowgill2017, kamiran2010, kamiran2012a, kamiran2012b, luong2011, kamishima2012, calders2010}), or otherwise change the distribution of one or more variables in the training data directly (perturbation; e.g. \cite{hajian2012, jiang2019identifying, wang2019repairing}). Referred to as data-massaging by \cite{zemel2013, kamiran2012a}, relabelling involves the modification of the labels of training data instances so that the proportion of positive instances are equal across all protected groups. It can also be applied to the test data upon the basis of strategies or probabilities learned on the training data. Often, but not always, approaches seek to retain the overall class distribution, i.e. the number of positive and negative instances remains the same. For example, \cite{luong2011} relabel the dependent variable (flip it from positive to negative or vice versa) if the data instance is determined as being discriminated against with respect to the observed outcome. Relabelling is also often part of counterfactual studies (e.g. \cite{heidari2018preventing, jung2018algorithmic, wang2019repairing}) which seek to investigate if flipping the dependent variable or other categorical sensitive variables affect the classification outcome.

Perturbation often aligns with notions of ``repairing'' some aspect(s) of the data with regard to notions of fairness. Applications in perturbation-based data repair \cite{hajian2012, kilbertus2019fair, jiang2019identifying, feldman2015, johndrow2019algorithm, gordaliza2019obtaining} have shown that accuracy is not significantly affected through perturbation. Whilst there are a number of papers that harness perturbation (it is not always referred to as perturbation) in the \ml literature, this approach appears more prevalent in the discrimination-aware data mining literature where it is often used as a means of privacy preservation. Often perturbation-based approaches are applied as a pre-processing step to prepare for an in-processing approach; often reweighing (e.g.\cite{jiang2019identifying}), and/or regularisation / optimization (e.g. \cite{wang2019repairing, ustun2019fairness}). It has also been proposed as a mechanism to detect proxy variables and variable influence \cite{wang2018avoiding} and counterfactual distributions \cite{wang2019repairing}.

Closely related to perturbation as a means to ``repair'' data is the use of sensitivity analysis (see \cite{saltelli2004sensitivity}) to explore how various aspects of the feature vector affect a given outcome. This is a relatively under-addressed area in the fairness literature, yet it has been well motivated (although perhaps indirectly): \cite{goodman2016} called for a better understanding of bias stemming from uncertainty, and \cite{hardt2016} who stressed that assessment of data reliability is needed. The application of sensitivity analysis in \ml is often to measure the stability of a model \cite{raginsky2016}. Whilst a number of approaches exist to determine model stability \cite{briand2009similarity, rakhlin2005, kutin2002, shalev2010, bousquet2002, kearns1999}, it has rarely been applied to \ml research beyond notions of model convergence and traditional performance measures with similar objectives to cross-validation. Yet, relabelling and perturbation are not far from the principals of sensitivity analysis. \cite{cowgill2017} proposed the perturbation of feature vectors to measure the effect on model performance of specific interventions. \cite{mckeown2013, gao2015} investigated visual mechanisms to better display ``issues'' with data to users, yet these approaches generally lack support for novice users \cite{berendt2014}. \cite{jung2018algorithmic, jung2017simple} used sensitivity analysis to evaluate sensitive variables and their relationship(s) with classification outcomes. Whilst sensitivity analysis is not a method to improve fairness (thus omitted from \figurename s \ref{fig:pre}-\ref{fig:postproc}) it can help to better understand uncertainty with respect to fairness. 

As for transformation, modification of the data via relabelling and perturbation is not always legally permissible \cite{barocas2016} and changes to the data should be minimised \cite{luong2011, hajian2012}. \cite{krasanakis2018adaptive} also note that some classifiers may be unaffected by the presence or specific nuances of some biases, and others may be negatively affected by altering the training data in an attempt to mitigate them. Thus, it is important to continuously (re)assess any fairness and methodological decisions made.
\subsection{Reweighing}
\label{sec:reweighing}
Unlike transformation, relabelling, and perturbation approaches which alter (certain instances of) the data, reweighing assigns weights to instances of the training data while leaving the data itself unchanged. Weights can be introduced for multiple purposes: 1) to indicate a frequency count for an instance type (e.g. \cite{calders2010}), 2) to place lower/higher importance on ``sensitive'' training samples (e.g. \cite{Calders2013, kamiran2012a, jiang2019identifying}), or 3) to improve classifier stability (e.g. \cite{krasanakis2018adaptive}). Reweighing as an approach straddles the boundary between pre-processing and in-processing. For example, \cite{kamiran2012a} seek to assign weights that take into consideration the likelihood of an instance with a specific class and sensitive value pairing (pre-processing). Whereas, \cite{krasanakis2018adaptive} first build an unweighted classifier, learn the weights of samples, then retrain their classifier using these weights (in-processing). A similar approach is taken by \cite{jiang2019identifying} who identify sensitive training instances (pre-processing), but then learn weights for these instances (in-processing) to optimize for the chosen fairness metric. 
 
With appropriate sampling (see \autoref{sec:sampling}), reweighing can maintain high(er) accuracy when compared to relabelling and blinding (omission) approaches \cite{kamiran2012a}. However, as \cite{krasanakis2018adaptive, globerson2006} note, classifier stability and robustness can be an issue. Thus \ml researchers need to carefully consider how reweighing approaches are applied and appropriately check for model stability. Reweighing also subtly changes the data composition making the process less transparent \cite{lepri2018, lum2016}.

\subsection{Regularization and Constraint Optimisation}
\label{sec:regularisation}

Classically, regularization is used in \ml to penalize the complexity of the learned hypothesis seeking to inhibit overfitting. When applied to fairness, regularization methods add one or more penalty terms which penalize the classifier for discriminatory practices \cite{kamishima2012}. Thus, it is not hypothesis (or learned model) driven, but data driven  \cite{bechavod2017penalizing} and based upon the notion(s) of fairness  considered. Much of the literature extends or augments the (convex) loss function of the classifier with fairness terms usually trying to find a balance between fairness and accuracy (e.g. \cite{kamishima2012, goel2018non, jiang2019wasserstein, heidari2018fairness, calders2010, Celis2019, manisha2018neural, zhang2018, berk2017, aghaei2019learning, feldman2015}). Some notable exceptions are  approaches which emphasise empirical risk subject to fairness constraints or welfare conditions (e.g. \cite{donini2018empirical, heidari2018preventing}), \tpr /\fpr of protected groups (e.g. \cite{bechavod2017penalizing}), stability of fairness (e.g. \cite{huang2019stable}), or counterfactual terms (e.g. \cite{di2020counterfactual}). 

\cite{huang2019stable, friedler2019comparative} note that often approaches for fair \ml are not stable, i.e. subtle changes in the training data significantly affect performance (comparatively high standard deviation). \cite{huang2019stable} argue that stability of fairness can be addressed through regularization and present corresponding empirical evidence through extensions of \cite{kamishima2012, zafar2017}. Aside from this, \cite{zemel2013, globerson2006} note that regularization as a mechanism is fairly generic, and can lead to a lack of model robustness and generalizability.

In-processing (constraint) optimization approaches (e.g. \cite{agarwal2018, Celis2019, chierichetti2019matroids, Haas2019, manisha2018neural, nabi2019learning, narasimhan2018learning, zemel2013, goh2016satisfying, nabi2018fair, cotter2019optimization, zafar2017, kim2018fairness}) have similar objectives to fairness regularization approaches, and hence we present them together. Constraint optimization approaches often include notions of fairness\footnote{We also note that many constraint optimisation papers often define new notions of fairness.} in the classifier loss function operating on the confusion matrix during model training. \cite{nabi2019learning} also approached this via reinforcement learning. Yet, these approaches can also include other constraints, and/or reduce the problem to a cost-sensitive classification problem (e.g. \cite{goh2016satisfying, agarwal2018, narasimhan2018learning}). Similarly, a multi-fairness metric approach has been proposed by \cite{kim2018fairness} where adaptions to stochastic gradient descent optimize weighted fairness constraints as an in-processing, or post-processing (when an pre-trained classifier is used) scenario.  \cite{narasimhan2018learning, goh2016satisfying, cotter2019optimization} summarizes a number of additional constraints types as precision or budget constraints to address the accuracy fairness trade off (often expressed as utility or risk functions e.g. \cite{corbett2017, Haas2019}); quantification or coverage constraints to capture disparities in class or population frequencies; churn constraints capturing online learning scenarios and enforcing that classifiers do not differ significantly from their original form as defined by the initial training data; and, stability constraints akin to the observations of \cite{huang2019stable, friedler2019comparative}. 

Key challenges for regularization approaches are: 1) they are often non-convex in nature or achieve convexity at the cost of probabilistic interpretation \cite{goel2018non}; 2) not all fairness measures are equally affected by the strength of regularization parameters \cite{di2020counterfactual, berk2017}; and 3) different regularization terms and penalties have diverse results on different data sets, i.e. this choice can have qualitative effects on the trade-off between accuracy and fairness \cite{berk2017}. For constraint optimization it can be difficult to balance conflicting constraints leading to more difficult or unstable training \cite{cotter2019optimization}.
\subsection{Adversarial Learning}
\label{sec:adversarial}
In adversarial learning the objective is for an adversary to try and determine whether a model training algorithm is robust enough. The framework of \cite{goodfellow2014generative} helped popularize the approach through the process of detecting falsified data samples \cite{celis2019improved}. When applied to applications of fairness in \ml, an adversary instead seeks to determine whether the training process is fair, and when not, feedback from the adversary is used to improve the model \cite{celis2019improved}. Most approaches in this area use notions of fairness within the adversary to apply feedback for model tuning as a form of in-processing where the adversary penalizes the model if a sensitive variable is predictable from the dependent variable (e.g. \cite{wadsworth2018achieving, beutel2017, zhang2018, edwards2015, xu2019achieving, celis2019improved, beutel2019putting}). This is often formulated as a multi-constraint optimization problem considering many of the constraints as discussed in \autoref{sec:regularisation}. There has also been work proposing the use of an adversary as a pre-processing transformation process on the training data (e.g. \cite{xu2018fairgan, madras2018, feng2019learning, xu2019achieving, adel2019one, kairouz2019censored}) with similar objectives to transformation as discussed in \autoref{sec:transformation} yet often moving towards a notion of ``censoring'' the training data with similar objectives to variable blinding as discussed in \autoref{sec:blinding}. Work has also started applying the notions of causal and counterfactual fairness to adversarial learning (e.g. \cite{xu2019achieving}). Here, the causal properties of the data prior to and after intervention are modeled with the adversarial intention to optimize a set of fairness constraints towards improved interventions.

An advantage of adversarial approaches is that they can consider multiple fairness constraints \cite{wadsworth2018achieving}, often treating the model as a blackbox \cite{madras2018}. However, adversarial approaches have been reported to often lack stability which can make them hard to train reliably \cite{beutel2019putting, feng2019learning} and also specifically in some transfer learning scenarios \cite{madras2018} when, for example, the protected variable is known only for a small number of samples. Additional forms of regularization have been proposed to try and address these issues (e.g. \cite{beutel2019putting}). The use of generative adversarial networks (GAN) with fairness considerations also permit applications within unstructured (for example multimedia) data or more generally as a generative process of creating an ``unbiased'' dataset using a number of samples. \cite{sattigeri2019fairness} illustrate this by using a GAN with fairness constraints to produce ``unbiased'' image datasets and \cite{xu2019achieving} have evidenced similar results for structured data.

\subsection{Bandits}
The application of bandits to \ml fairness \cite{ensign2018decision, joseph2016fairness, joseph2018meritocratic, liu2017calibrated, gillen2018online} is nascent. As yet, papers have proofs of their work but lack general evaluation against specific data. As a reinforcement learning framework, bandits are motivated on the need for decisions to be made in an online manner \cite{joseph2016fair}, and that decision makers may not be able to define what it means to be ``fair'' but that they may recognize ``unfairness'' when they see it \cite{gillen2018online}. Approaches that use bandits do so on the basis of \cite{dwork2012}'s notion of individual fairness, i.e. that all similar individuals should be treated similarly. Thus, bandit approaches frame the fairness problem as a stochastic multi-armed bandit framework, assigning either individuals to arms, or groups of ``similar'' individuals to arms, and fairness quality as a reward represented as regret \cite{joseph2016fair, liu2017calibrated}. The two main notions of fairness that have emerged from the application of bandits are meritocratic fairness \cite{joseph2016fair, joseph2018meritocratic} (group agnostic) and subjective fairness \cite{liu2017calibrated} (emphasises fairness in each time period $t$ of the bandit framework).

\subsection{Calibration}
\label{sec:calibration}
Calibration is the process of ensuring that the proportion of positive predictions is equal to the proportion of positive examples \cite{dawid1982well}. In the context of fairness, this should also hold for all subgroups (protected or otherwise) in the data \cite{pleiss2017, zhang2016identifying, chen2018discriminatory}. Calibration is particularly useful when the output is not a direct decision but used to inform human judgement when assessing risks (e.g. awarding a loan) \cite{noriega2019active}. As such, a calibrated model does not inhibit biases of decision makers, but rather ensure that risk estimates for various (protected) subgroups carry the same meaning \cite{pleiss2017}. However calibrating for multiple protected groups and/or using multiple fairness criteria at once has been shown to be impossible \cite{chouldechova2017, kleinberg2016inherent,liu2019implicit, hebert2017calibration, liu2017calibrated, pleiss2017, noriega2019active, kim2018fairness}. \cite{pleiss2017} even note that the goals of low error and calibration are competing objectives for a model.  This occurs as calibration has limited flexibility \cite{jiang2019identifying}. \cite{woodworth2017} also evidenced that decoupling the classifier training from the means to increase fairness, i.e. post-processing, is provably sub-optimal. 

The literature has proposed various approaches to handle the impasse of achieving calibration and other fairness measures. One approach has been to apply a randomization post-processing process to try and achieve a balance between accuracy and fairness, yet \cite{kleinberg2016inherent, corbett2018, hardt2016, pleiss2017, noriega2019active} discuss a number of shortcomings of this approach. Notably, the individuals who are randomized are not necessarily positively impacted, and the overall accuracy of the model can be adversely affected. \cite{noriega2019active} also note that this approach is Pareto sub-optimal, and instead propose a cost-based approach to balance calibration and error parity. \cite{liu2019implicit} suggest that calibration is sufficient as a fairness criterion if the model is unconstrained. \cite{hebert2017calibration} instead seek to achieve approximate calibration (i.e. to guarantee calibration with high probability) using a multi-calibration approach that operates on identifiable subgroups to balance individual and group fairness measures even for small samples: a specific challenge for achieving calibration \cite{liu2019implicit}. \cite{kearns2018preventing, kim2019multiaccuracy} undertake a similar approach under different settings.  \cite{liu2017calibrated} have proposed a bandit-based approach to calibration. 

\subsection{Thresholding}
\label{sec:thresholding}

Thresholding is a post-processing approach which is motivated on the basis that discriminatory decisions are often made close to decision making boundaries because of a decision maker's bias \cite{kamiran2012b} and that humans apply threshold rules when making decisions \cite{kleinberg2018human}. Thresholding approaches often seek to find regions of the posterior probability distribution of a classifier where favored and protected groups are both positively and negatively classified.\footnote{We note that there is a fine line between thresholding and calibration approaches and that they often overlap.} Such instances are considered to be ambiguous, and therefore potentially influenced by bias \cite{kamiran2012b}. To handle this, researchers have devised approaches to determine threshold values via measures such as equalized odds specifically for different protected groups to find a balance between the true and false positive rates to minimize the expected classifier loss \cite{hardt2016}. The underlying idea here is to incentivize good performance (in terms of both fairness and accuracy) across all classes and groups. 

Computing the threshold value(s) can be undertaken either by hand to enable a user to assert preferences with respect to the fairness accuracy trade-off or use other statistical methods. \cite{menon2017cost} estimate the thresholds for each protected group using logistic regression, then use a fairness frontier to illustrate disalignment between threshold values. \cite{kamiran2012b} use an ensemble to identify instances in an uncertainty region and assist in setting a threshold value. \cite{fish2016confidence} propose a method to shift decision boundaries using a form of post-processing regularization. \cite{valera2019enhancing} use posterior sampling to maximize a fairness utility measure. \cite{iosifidis2020fae} learn a threshold value after training an ensemble of decoupled ensembles (a pre-processing sampling approach, see \autoref{sec:sampling}) such that the discrepancy between protected and non-protected groups is less than some user specified discrimination threshold value. Thus, one of the key challenges for thresholding approaches is to determine preferences with respect to a tolerance for unfairness. \cite{iosifidis2020fae} note that this is often undertaken with respect to accuracy, but in many cases class imbalance would invalidate such decisions. Thresholding is often argued as a potential human-in-the-loop mechanism, yet in the absence of appropriate training programs (for the human-in-the-loop) this can introduce new issues \cite{veale2017}. This stems from fairness typically not being representable as a monotone function, therefore assigning a threshold value may be quite arbitrary \cite{speicher2018}. Thresholding approaches often claim compelling notions of equity, however, only when the threshold is correctly chosen \cite{corbett2018}.



\section{Beyond Binary Classification}
\label{sec:beyond}

The bulk of the fairness literature focuses on binary classification  \cite{berk2017}. In this section, we provide an overview and discussion beyond approaches for binary classification (albeit less comprehensive) and note that there is a sufficient need for fairness researchers to also focus on other \ml problems. 

\subsection{Fair Regression}
\label{sec:regression}
\simon{add: \cite{heidari2018preventing} (post-processing fair regression with lasso)?}

The main goal of fair regression is to minimize a loss function $l(Y, \hat{Y})$, which measures the difference between actual and predicted values, while also aiming to guarantee fairness. The general formulation is similar to the case of (binary) classification, with the difference that $Y$ and $\hat{Y}$ are continuous rather than binary or categorical. 
Fairness definitions for regressions adapt principles defined in  \autoref{sec:metrics}. For example, parity-based metrics aim to make the loss function equal for different groups \cite{agarwal2019fair}. 
With respect to defining fairness metrics or measurements, \cite{berk2018fairness} suggest several metrics that can be used for general regression models. \cite{agarwal2019fair} define both statistical parity and bounded-group-loss metrics to measure fairness in regression, the latter providing a customizable maximum allowable loss variable that defines a specific trade-off between fairness and loss (and thus predictor performance). 
\cite{calders2013fairlinear} consider biases in linear regression as measured by the effects of a sensitive attribute on $Y$ through the mean difference (difference of mean target variable between groups) and AUC metrics. They suggest the use of propensity modeling as well as additional constraints (e.g., enforcing a mean difference of zero) to mitigate biases in linear regression. 

The effect of bias on parameter estimates and coefficients in multiple linear regression is discussed by \cite{johnson2016impartial}, who also suggest a post-processing approach to make parameter estimates impartial with respect to a sensitive attribute. \cite{komiyama18a-plmr} include fairness perspectives in non-convex optimization for (linear) regression using the coefficient of determination between the predictions $\hat{y}$ and the sensitive attribute(s) as additional constraints in their (constrained) linear least squares model that generates a solution for a user-selected maximum level for the coefficient of determination. 

\cite{perez2017fair} propose methods for fair regression as well as fair dimensionality reduction using a Hilbert Schmidt independence criterion and a projection-based methodology that is able to consider multiple sensitive attributes simultaneously. \cite{kamishima2012} suggest a regularization approach that can be applied to general prediction algorithms. \cite{fukuchi2013} define the concept of $\mu$-neutrality that measures if probabilistic models are neutral with respect to specific variables and show that this definition is equivalent to statistical parity. 
\cite{berk2017} propose a family of regularization approaches for fair regression that works with a variety of group and individual fairness metrics. Through a regularization weight, the proposed method is able to calculate accuracy-fairness trade-offs and evaluate the efficient frontier of achievable accuracy-fairness trade-offs. 
\cite{fitzsimons2019general} consider group-based fairness metrics and their inclusion in kernel regression methods such as decision tree regression while keeping efficiency in computation and memory requirements.

\subsection{Recommender Systems}
\label{sec:recsys}


Considerations of fairness have been actively studied in the context of rankings and recommender systems. 
For rankings in general, \cite{Yang2017}, \cite{zehlike2017}, and \cite{biega2018} define different types of fairness notions such as group-based fairness in top-k ranking (\cite{Yang2017,zehlike2017}), an individual fairness measure in rankings following concepts similar to \cite{hardt2016} and the equality of opportunity in binary classification  \cite{biega2018}, and unfairness of rankings over time through a dynamic measure called amortized fairness \cite{biega2018}. 

For recommender systems in particular, \cite{Lee2014} consider fairness-aware loan recommendation systems and argue that fairness and recommendation are two contradicting tasks. They measure fairness as the standard deviation of the top-N recommendations, where a low standard deviation signifies a fair recommendation without compromising accuracy. Subsequent publications expanded this view of recommender fairness by proposing additional metrics as well as algorithms \cite{Yao2017,zhu2018fairness,steck2018,beutel2019,Chakraborty2019}. These include a set of \ml inspired group-based fairness metrics that address different forms of unfairness to address potential biases in collaborative filtering recommender systems stemming from a population imbalance or observation bias \cite{Yao2017}, fairness goals for recommender systems as overcoming algorithmic bias and making neutral recommendations independent of group membership (e.g., based on gender or age) \cite{zhu2018fairness}, recommendation calibration, i.e., the proportional representation of items in recommendations, \cite{steck2018}, pairwise fairness as well as a regularization approach to improve model performance \cite{beutel2019}, and two fairness measures in top-k recommendations, proportional representation and anti-plurality \cite{Chakraborty2019}. Further approaches include tensor-based recommendations have been proposed that take statistical parity into account \cite{zhu2018fairness}, and a mechanism design approach for fairly dividing a set of goods between groups using disparate impact as fairness measure and a recommender system as evaluation use case \cite{peysakhovich2019fair}.

An aspect that sets fairness considerations in recommender systems apart from binary classification in \ml is that fairness can be seen as multi-sided concept that can be relevant for both users (who consume the recommendations) and items. \cite{Burke2017,pmlr-v81-burke18a} introduce the notion of ``C-fairness'' for fair user/consumer recommendation (user-based), and ``P-fairness'' for fairness of producer recommendation (item-based) to address this multi-sided aspect, showing that defining generalized approaches to multi-sided fairness is hard due to the domain specificity of the multi-stakeholder environment. \cite{Ekstrand2018} presents an empirical analysis of P-fairness for several collaborative filtering algorithms. 
Similarly, \cite{zheng2018fairness} aim to find an optimal trade-off between the utilities of the multiple stakeholders. Other work considering fairness from either the consumer or provider side include the analysis of different recommendation strategies for a variety of (fairness) metrics \cite{Jannach2015}, subset-based evaluation metrics to measure the utility of recommendations for different groups (e.g., based on demographics) \cite{pmlr-v81-ekstrand18b}, and a general framework to optimize utility metrics under fairness of exposure constraints \cite{singh2018,singh2019}. 
Besides the previous work on metrics and algorithms, several authors have also proposed bias mitigation strategies. This includes a pre-processing approach (recommendation independence) to make recommendations independent of a specific attribute \cite{pmlr-v81-kamishima18a}, adding specifically designed ``antidote'' data to the input instead of manipulating the actual input data to improve the social desirability of the calculated recommendations \cite{Rastegarpanah2019}, and a post-processing algorithm to improve user-based fairness through calibrated recommendations \cite{steck2018}.

\subsection{Unsupervised Methods}
\label{sec:unsupervised}
Currently, unsupervised methods fall into three distinct areas: 1) fair clustering (e.g. \cite{schmidt2019fair, chierichetti2017, bera2019fair, rosner2018privacy, backurs2019scalable, bercea2018cost, bera2019fair, chen2019proportionally}); 2) investigating the presence and detection of discrimination in association rule mining (e.g. \cite{pedreshi2008, pedreschi2012study, hajian2012}); and 3) transfer learning (e.g. \cite{dwork2018, coston2019fair}).

Fair clustering started with the initial work of \cite{chierichetti2017} who introduced the idea of micro-cluster fairlet decomposition as a pre-processing stage applied prior to standard centroid-based methods like k-means and k-medians. Thus far, clustering approaches have mostly operating on \cite{feldman2015}'s introduction of disparate impact introducing this as cluster balance, where balance pertains to uniformity of distribution over $k$ clusters of belonging to some protected group. \cite{chierichetti2017} use color to represent belonging to the protected group or not. When multiple protected groups are in place this means optimising for both the number of clusters, and the number as well as spread of colors. This is undertaken by  \cite{bera2019fair} who extend the work of \cite{chierichetti2017} to allow for more than two colors and fuzzy cluster membership functions arguing that otherwise the approach is too stringent and brittle. Yet, there is a cost here, unlike other approaches to fairness in \ml , fair clustering has significant computational costs associated to it. However,  methods have emerged to handle this via coresets \cite{schmidt2019fair} and approximate fairlet decomposition \cite{backurs2019scalable}. Fair clustering has also seen applications in the the discovery of potentially protected groups (e.g. \cite{benthall2019racial, chierichetti2019matroids}).

Approaches that utilize transfer learning do so in combination with other methods. The motivation for using transfer learning is typically in response to an observable covariate shift between the source (training) and target distributions. This can often occur in real world application settings and requires that the model is trained on a different probability distribution to that which the model will ultimately be tested (and later deployed) on \cite{bickel2009discriminative,sugiyama2007covariate, quionero2009dataset}. Here, transfer learning acts as an unsupervised domain adaption technique to account for such covariate shifts \cite{gong2012geodesic, torrey2010transfer, dwork2018}. In this, transfer learning approaches are somewhat analogous to reweighing approaches in that they seek to determine weights for each training example that account for a covariate shift optimized using regularization techniques (e.g. \cite{coston2019fair}) or forms of joint loss functions (e.g. \cite{dwork2018}). 
\subsection{Natural Language Processing}
\label{sec:nlp}

Natural language processing (NLP) is an area of machine learning which operates on text data, e.g. document classification, sentiment analysis, text generation, translation, etc. Unintended biases have also been noticed in NLP; these are often gender or race focused \cite{Homemaker, park2018reducing, diaz2018addressing, dixon2018measuring, kiritchenko2018examining, zhiltsova2019mitigation, Caliskan2017, zhiltsova2019mitigation}. \cite{Homemaker} highlighted that word embeddings (often used for various tasks ranging from Google news sorting to abusive content filtering) can be biased against women. \cite{kiritchenko2018examining} notes that sentiment analysis systems can discriminate against races and genders noting that specific race or gender references in text can result negative sentiment whereas different race or gender references the same text is noted as positive. Unintended bias can be introduced to the data by personal biases during manual data labelling \cite{dixon2018measuring} as well as biases that occur in language use for example through non-native speaker errors \cite{zhiltsova2019mitigation, Tatman, blodgett2017racial} or how the text data is prepared as well as the general model architecture  \cite{park2018reducing}. In terms of approaches, combating biases in NLP occur mostly in the pre-processing stage, such as removing or replacing specific words (e.g. \cite{diaz2018addressing}), dictionary modification (e.g. \cite{zhiltsova2019mitigation}), and unsupervised balancing of the training dataset \cite{dixon2018measuring}. However, in some cases mitigation of biases is not attempted due to the complexity of the task \cite{zhiltsova2019mitigation, kiritchenko2018examining}.

\section{Current Platforms}
\label{sec:platforms}

While many researchers publish their individual approaches on github and similar platforms, a few notable projects have emerged that address fairness in \ml from a more general perspective. \autoref{tbl:platformoverview} describes some of these approaches. We note the existence of proprietary software, yet here emphasize readily tools to discuss the current state of the art for \ml researchers and practitioners.

\begin{table}[h]
    \scriptsize
    \centering
    \begin{tabularx}{\textwidth}{p{.2\textwidth}@{\hskip 0.15in}X}
        \toprule
         Project &  Features   \\
        \midrule
        AIF360 \cite{bellamy2018} &  Set of tools that provides several pre-, in-, and post-processing approaches for binary classification as well as several pre-implemented datasets that are commonly used in Fairness research 
        \\
        Fairlean\footnotemark & Implements several parity-based fairness measures and algorithms \cite{hardt2016,agarwal2018,agarwal2019fair} for binary classification and regression as well as a dashboard to visualize disparity in accuracy and parity. 
        \\
        Aequitas \cite{saleiro2018aequitas}  &  Open source bias audit toolkit. Focuses on standard \ml metrics and their evaluation for different subgroups of a protective attribute.
        \\
        Responsibly \cite{louppe2016learning}  & Provides datasets, metrics, and algorithms to measure and mitigate bias in classification as well as NLP (bias in word embeddings). 
        \\
        Fairness\footnotemark &  Tool that provides commonly used fairness metrics (e.g., statistical parity, equalized odds) for R projects.  
        \\
        
        FairTest \cite{tramer2017fairtest}  & Generic framework that provides measures and statistical tests to detect unwanted associations between the output of an algorithm and a sensitive attribute.  
        \\
        Fairness Measures\footnotemark &  Project that considers quantitative definitions of discrimination in classification and ranking scenarios. Provides datasets, measures, and algorithms (for ranking) that investigate fairness.  
        \\ 
        Audit AI\footnotemark & Implements various statistical significance tests to detect discrimination between groups and bias from standard machine learning procedures.  
        \\
        Dataset Nutrition Label \cite{holland2018dataset} & Generates qualitative and quantitative measures and descriptions of dataset health to assess the quality of a dataset used for training and building \ml models. 
        \\
        ML Fairness Gym  & Part of Google's Open AI project, a simulation toolkit to study long-run impacts of ML decisions.\footnotemark   Analyzes how algorithms that take fairness into consideration change the underlying data (previous classifications) over time (see e.g. \cite{liu2018plmr,elzayn2019fair,ensign2018runaway,hu2019manipulation,milli2019social}). 
        \\
        \bottomrule        
    \end{tabularx}
    \caption{Overview of projects addressing Fairness in Machine Learning.}
    \label{tbl:platformoverview}
    \tightentab
\end{table}

\addtocounter{footnote}{-5}

\footnotetext{\url{https://github.com/fairlearn/fairlearn}}
\addtocounter{footnote}{1}
\footnotetext{\url{https://github.com/kozodoi/Fairness}}
\addtocounter{footnote}{1}
\footnotetext{\url{https://github.com/google-research/tensorflow_constrained_optimization}}
\addtocounter{footnote}{1}
\footnotetext{\url{https://github.com/pymetrics/audit-ai}}
\addtocounter{footnote}{1}
\footnotetext{\url{http://www.fairness-measures.org/}, \url{https://github.com/megantosh/fairness_measures_code/tree/master}}
\addtocounter{footnote}{1}
\footnotetext{\url{https://github.com/google/ml-fairness-gym}} 
\addtocounter{footnote}{1}

\section{Concluding Remarks: The Fairness Dilemmas}
\label{sec:dilemmas}
\simon{the data science dilemma: performance vs. fairness vs. transparency vs. complexity vs. competence}

In this article, we have provided an introduction to the domain of fairness in Machine Learning (ML) research. This encompasses a general introduction (\autoref{sec:overview}), different measures of fairness for \ml (\autoref{sec:metrics}), methods to mitigate bias and unfairness in binary classification problems (\autoref{sec:approaches}) as well as beyond binary classification (\autoref{sec:beyond}) and listed some open source tools (\autoref{sec:platforms}) to assist researchers and practitioners seeking to enter this domain or employ state of the art methods within their \ml pipelines. 
For specific methods, we have noted the key challenges of their deployment. Now, we focus on the more general challenges for the domain as a whole as a set of four dilemmas for future research (the ordering is coincidental and not indicative of importance): {\bf Dilemma-1:} Balancing the trade-off between fairness and model performance (\autoref{sec:tradeoff}); {\bf Dilemma-2:} Quantitative notions of fairness permit model optimization, yet cannot balance different notions of fairness, i.e individual vs. group fairness (\autoref{sec:balance}); {\bf Dilemma-3:} Tensions between fairness,  situational, ethical, and sociocultural context, and policy (\autoref{sec:context}); and {\bf Dilemma-4:} Recent advances to the state of the art have increased the skills gap inhibiting ``man-on-the-street'' and industry uptake (\autoref{sec:skills}).

\subsection{Dilemma 1: Fairness vs. Model Performance}
\label{sec:tradeoff}
A lack of consideration for the sociocultural context of the application can result in \ml solutions that are biased, unethical, unfair, and often not legally permissible \cite{boddington2017, yeung2018}. The \ml community has responded with a variety of mechanisms to improve the fairness of models as outlined in this article. However, when implementing fairness measures, we must emphasize either fairness or model performance as improving one can often detriment the other \cite{berk2018fairness, dwork2012, corbett2017, hardt2016, zliobaite2015, calmon2017, Haas2019}. \cite{feldman2015} do, however, note that a reduction in accuracy may in fact be the desired result, if it was discrimination in the first place that raised accuracy. Note that even prior to recognizing this trade-off, we need to be cautious in our definition of model performance. \ml practitioners can measure performance in a multitude of ways, and there has been much discussion concerning the choice of different performance measures and approaches \cite{demvsar2006statistical, sokolova2006beyond}. The choice of performance measure(s) itself may even harbor, disguise, or create new underlying ethical concerns. We also note that currently, there is little runtime benchmarking of methods outside of clustering approaches (see \cite{schmidt2019fair, backurs2019scalable}). This is an observation as opposed to a criticism, but we note that potential users of fairness methods will likely concern themselves with computational costs, especially if they increase. 

\subsection{Dilemma 2: (Dis)agreement and Incompatibility of ``Fairness''}
\label{sec:balance}
On top of the performance trade-off, there is no consensus in literature whether individual or group fairness should be prioritized. Fairness metrics usually either emphasize individual or group fairness, but cannot combine both  \cite{kleinberg2018, chouldechova2017}. \cite{speicher2018} also note that many approaches to group fairness often tackle between-group issues, as a consequence they demonstrate that within-group issues are worsened through this choice. To further complicate things, \cite{corbett2018, green2018} argue that with a reliance on expressing fairness mathematically these definitions often do not map to normative social, economic, or legal understandings of the same concepts. This is corroborated by \cite{skirpan2017} who note an over-emphasis in the literature on specific measures of fairness and insufficient dialogue between researchers and affected communities. Thus, improving fairness in \ml is challenging and simultaneously there are many different notions for researchers and practitioners to navigate. Further adding to this discussion is the notion of differing views of the root(s) of fairness and bias. \cite{pierson2017gender, pierson2017demographics, grgic2018human, srivastava2019mathematical} study the differing views of people in this regard and observe that this is not a trivial challenge to address. E.g., \cite{pierson2017gender} notes that women have differing views in the inclusion / exclusion of gender as a protected variable to men. \cite{Hutchinson2019a} note that a similar discussion was left unresolved in the early days of fairness research in the context of test scores and employment/hiring practices, indicating that this is one of the main challenges of \ml fairness research in the future. \cite{kallus2018residual} have noted that this dilemma can be articulated as a bias in, bias out property of \ml: i.e. addressing one form of bias results in another. 

Thus, the community as articulated in \cite{chen2018discriminatory, hebert2017calibration, buolamwini2018gender} needs to explore ways to either handle combinations of fairness metrics, even if only approximately due to specific incompatibilities, or implement a significant meta review of measures to help categorise specific differences, ideological trade-offs, and preferences. This will enable researchers and practitioners to consider a balance of the fairness measures they are using. This is a challenging undertaking and whilst the tools discussed in \autoref{sec:platforms} go some way to facilitate this, there is a need for more general toolkits and methodologies for comparing fairness approaches. We note a number of comparative studies, i.e. \cite{friedler2019comparative, galhotra2017fairness, tramer2017fairtest}, but these only scratch the surface. 

\subsection{Dilemma 3: Tensions with Context and Policy}
\label{sec:context}
The literature typically hints toward ``optimizing'' fairness without transparency of the root(s) of (un)fairness \cite{lepri2018} rarely extending beyond ``(un)fair'' \cite{cowgill2017, speicher2018} typically to mirror current legal thought \cite{feldman2015}. This is true for both metrics and methods. As such, platforms are needed to assist practitioners in ascertaining the cause(s) of unfairness and bias. However, beyond this, critics of current research \cite{burrell2016, skirpan2017, veale2017, veale2018, lepri2017, yeung2018, wong2019democratizing, wong2019democratizing} argue that efforts will fail unless contextual, sociocultural, and social policy challenges are better understood. Thus, there is an argument that instead of striving  to ``minimize'' unfairness, more awareness of context-based aspects of discrimination is needed. There is the prevalent assumption that ``unfairness'' has a uniform context-agnostic egalitarian valuation function for decision makers when considering different (sub)populations \cite{binns2017, corbett2017, corbett2018}. This suggests a disconnect between organizational realities and current research, which undermines advancements \cite{veale2018, lipton2016}. Other suggestions have been for \ml researchers and practitioners to better understand the limitations of human decision making \cite{rich2019lessons}.  

It is easy to criticize, however, the underlying challenge is availability of realistic data. Currently, the literature relies unilaterally on convenience datasets (enabling comparative studies), often from the UCI repository \cite{uci} or similar with limited industry context and engagement  \cite{veale2018, veale2017, lepri2017}. \cite{kilbertus2019fair, woodworth2017, lakkaraju2017learning,lakkaraju2017selective, kallus2018balanced} note that there is an additional challenge in the datasets used to train models: data represent past decisions, and as such inherent bias(es) in these decisions are amplified. This is a problem referred to as selective labels \cite{lakkaraju2017selective}. Similarly, there may be differences in the distribution(s) of the data between the data the model is trained on, and deployed on: dataset shift as discussed by \cite{quionero2009dataset}. As such, data context cannot be disregarded. 

Thus, researchers need to better engage with (industry) stakeholders to study models in vivo and engage proactively in open debate on policy and standardization. This is a hard problem to solve: companies cannot simply hand out data to researchers and researchers cannot fix this problem on their own. There is a tension here between advancing the fairness state of the art, privacy \cite{sokolovska2018integrating, fung2010privacy}, and policy. \cite{veale2018} notes that policy makers are generally not considered or involved in the \ml fairness domain. We are seeing an increasing number of working groups on best practices for ethics, bias, and fairness, where Ireland's NSAI/TC 002/SC 18 Artificial Intelligence working group, the IEEE P7003 standardization working group on algorithmic bias, and the Big Data Value Association are just three examples of many, but this needs to be pushed harder at national and international levels by funding agencies, policy makers, and researchers themselves.




\subsection{Dilemma 4: Democratisation of ML vs the Fairness Skills Gap}
\label{sec:skills}
Today, \ml technologies are more accessible than ever. This has occurred through a combination of surge in third level courses and the wide availability of \ml tools such as WEKA \cite{hall2009weka}, RapidMiner\footnote{\url{https://rapidminer.com}}, and SPSS Modeler\footnote{\url{https://www.ibm.com/ie-en/products/spss-modeler}}. Alternatively, Cloud-based solutions such as Google's Cloud AutoML \cite{bisong2019google}, Uber AI's Ludwig,\footnote{https://uber.github.io/ludwig/} and Baidu's EZDL\footnote{\url{https://ai.baidu.com/ezdl/}} remove the need to even run models locally. The no-/low-code \ml movement is arguably enabling more  companies to adopt \ml technologies.
In addition, there is a growing trend in the use of Automated Machine Learning (AutoML) \cite{thornton2013auto, feurer2015efficient} to train \ml models. AutoML abstracts much of the core methodological expertise (e.g., KDD \cite{fayyad1996kdd}, and CRISP-DM \cite{chapman2000crisp}) by automated feature extraction and training multiple models often combining them into an ensemble of models that maximizes a set of performance measures. Collectively, each of these advancements positively democratizes \ml, as it means lower barriers of use: ``push button operationalization'' \cite{team2016azureml} with online marketplaces\footnote{E.g.: Amazon's \url{https://aws.amazon.com/marketplace/solutions/machinelearning/} and Microsoft's \url{https://gallery.azure.ai} ML Marketplaces.} selling third party \ml solutions. 

Lowering the entry barrier to \ml through democratization will (if it hasn't already) mean an increase in (un)intentional socially insensitive uses of \ml technologies. The challenge is that \ml application development follows a traditional software development model: it is modular, sequential, and based on large collections of (often) open source libraries, but methods to highlight bias, fairness, or ethical issues assume high expertise in \ml development and do not consider ``on-the-street'' practitioners \cite{veale2018, lepri2018}. This was our motivation in writing this survey. However, the fairness domain is only just starting to provide open source tools available for practitioners (\autoref{sec:platforms}). Yet, in general there is little accommodation for varying levels of technical proficiency, and this undermines current advancement \cite{skirpan2017, goodman2016, veale2018, veale2017, burrell2016}. There is a tension between educational programs (as called for in \cite{burrell2016}) and the degree of proficiency needed to apply methods and methodologies for fair \ml. \cite{veale2017, corbett2018} have advocated this as the formalization of exploratory fairness analysis: similar to exploratory data analysis, yet for informed decision making with regard to ``fair'' methodological decisions. Similarly, \cite{saltz2019integrating} call for core \ml educational resources and courses to better include ethical reasoning and deliberation and provide an overview of potential materials. Thus, the fourth dilemma currently facing the fair \ml domain is its own democratization to keep up with the pace of \ml proliferation across sectors. This means a shift in terms of scientific reporting, open source comprehensive frameworks for repeatable and multi-stage (i.e., pipelined models) decision making processes where one model feeds into another \cite{bower2017, goodman2016, thomas2019preventing}. Currently under-addressed is bias and fairness transitivity: where one \ml model is downstream to another.


\subsection{Concluding Remarks}
The literature almost unilaterally focuses on supervised learning with an overwhelming emphasis on binary classification \cite{berk2017}: diversification is needed. With very few exceptions, the approaches discussed in this article operate on the assumption of some set of (usually a priori known) ``protected variables''. This doesn't help practitioners. Tools potentially based on causal methods (\autoref{sec:repair}) are needed to assist in the identification of protected variables and groups as well as their proxies. 

More realistic datasets are needed: \cite{raginsky2016} argue that approaches tend to operate on too small a subset of features raising stability concerns. This should go hand in hand with more industry-focused training. Tackling fairness from the perspective of protected variables or groups needs methodological care, as ``fixing'' one set of biases may inflate another \cite{binns2017, corbett2018} rendering the model as intrinsically discriminatory as a random model \cite{pleiss2017, dwork2018}.  There is also the risk of redlining, where although the sensitive attribute is ``handled'' sufficiently, correlated variables are still present \cite{pedreshi2008, romei2014, zarsky2016, veale2017, dwork2012, calmon2017}, amplifying instead of reducing unfairness \cite{dwork2012}. 

We also note specific considerations of pre-processing vs. in-processing vs. post-processing interventions. Pre-processing methods, which modify the training data, are at odds with policies like \gdpr 's right to an explanation, and can introduce new subjectivity biases \cite{veale2017}. They also assume sufficient knowledge of the data, and make assumptions over its veracity \cite{corbett2018}. Uptake of in-processing approaches requires better integration with standard \ml libraries to overcoming  porting challenges. \cite{woodworth2017} noted that generally post-processing methods have suboptimal accuracy compared to other ``equally fair'' classifiers, with \cite{agarwal2018} noting that often test-time access to protected attributes is needed, which may not be legally permissible, and have other undesirable effects \cite{chen2018discriminatory}.

 
As a closing thought many approaches to reduce discrimination may themselves be unethical or impractical in settings where model accuracy is critical such as in healthcare, or criminal justice scenarios \cite{chen2018discriminatory}. This is not to advocate that models in these scenarios should be permitted to knowingly discriminate, but rather that a more concerted effort is needed to understand the roots of discrimination.  Perhaps, as \cite{chen2018discriminatory, liu2018plmr, ensign2018runaway, woodworth2017} note, it may often be better to fix the underlying data sample (e.g. collect more data, which better represents minority or protected groups and delay the modeling phase) than try to fix a discriminatory \ml model.

\bibliographystyle{plainnat}
\bibliography{refs}

\end{document}